\documentclass{article}
\usepackage{hyperref}
\usepackage{microtype}
\usepackage{graphicx}
\usepackage{subfig}
\usepackage{booktabs,amsmath,amssymb} 
\usepackage[normalem]{ulem}
\useunder{\uline}{\ul}{}
\usepackage[table,xcdraw]{xcolor}
 \usepackage{multirow, float}
 \usepackage{hhline}
\usepackage{wrapfig, mathtools, url}
\usepackage{float}

\newcommand{\x}{\mathrm{x}}
\newcommand{\y}{\mathrm{y}}
\newcommand{\F}{\mathrm{F}}
\newcommand{\xs}{\mathcal{X}}
\newcommand{\ys}{\mathcal{Y}}
\newcommand{\alg}[0]{\texttt{SiSTA}}
\newcommand{\algg}[0]{\texttt{SiSTA-G}}
\newcommand{\alggmc}[0]{\texttt{SiSTA-mcG}}
\newcommand{\algs}[0]{\texttt{SiSTA-S}}
\newcommand{\jay}[1]{\textcolor{orange}{[Jay]}}
\newcommand{\rak}[1]{\textcolor{green}{[Rak]}}
\newcommand{\kow}[1]{\textcolor{red}{[Kow]}}

\usepackage[accepted]{icml2023}

\icmltitlerunning{Target-Aware Generative Augmentations for Single-Shot Adaptation\hfill\thepage}
\begin{document}

\twocolumn[
\icmltitle{Target-Aware Generative Augmentations for Single-Shot Adaptation}



\icmlsetsymbol{equal}{*}

\begin{icmlauthorlist}
\icmlauthor{Kowshik Thopalli}{equal,llnl}
\icmlauthor{Rakshith Subramanyam}{equal,asu}
\icmlauthor{Pavan Turaga}{asu}
\icmlauthor{Jayaraman J. Thiagarajan}{llnl}
\end{icmlauthorlist}

\icmlaffiliation{llnl}{Lawrence Livermore National Laboratory, Livermore, CA, USA}
\icmlaffiliation{asu}{Arizona State University, Tempe, AZ, USA}
\icmlcorrespondingauthor{Kowshik Thopalli}{thopalli1@llnl.gov}

\icmlkeywords{Domain Adaptation, StyleGAN, Data Augmentation, Deep Learning, Distribution Shifts}

\vskip 0.3in
]



\printAffiliationsAndNotice{\icmlEqualContribution} 

\begin{abstract}
In this paper, we address the problem of adapting models from a source domain to a target domain, a task that has become increasingly important due to the brittle generalization of deep neural networks. 
While several test-time adaptation techniques have emerged, they typically rely on synthetic toolbox data augmentations in cases of limited target data availability. We consider the challenging setting of single-shot adaptation and explore the design of augmentation strategies. We argue that augmentations utilized by existing methods are insufficient to handle large distribution shifts, and hence propose a new approach \alg~(\underline{Si}ngle-\underline{S}hot \underline{T}arget \underline{A}ugmentations), which first fine-tunes a generative model from the source domain using a single-shot target, and then employs novel sampling strategies for curating synthetic target data. Using experiments on a variety of benchmarks, distribution shifts and image corruptions, we find that \alg~produces significantly improved generalization over existing baselines in face attribute detection and multi-class object recognition. Furthermore, \alg~performs competitively to models obtained by training on larger target datasets.
Our codes can be accessed at \url{https://github.com/Rakshith-2905/SiSTA}.
\end{abstract}

\section{Introduction}
\label{sec:intro}

Deep models tend to suffer a significant drop in their performance when there is a shift between train and test distributions~\cite{torralba2011unbiased}. A natural solution to improve generalization under such domain shifts is to adapt models using data from the target domain of interest. However, it is infeasible to obtain data from every possible target during source model training itself. Test-time adaptation has emerged as an alternate solution, where a source-trained model is adapted solely using target data without accessing the source data. However, the success of these source-free adaptation (SFDA) methods hinges on sufficient target data availability~\cite{SHOT,yang2021exploiting}. While there exist online adaptation methods such as TENT~\cite{Tent} and MEMO~\cite{zhang2021memo}, they are are found to be ineffective under complex distribution shifts and when target data is limited, often producing on par or only marginally better results than non-adaptation performance~\cite{cattan}.

In this work, we investigate a practical, yet challenging, scenario where the goal is to adapt models under unknown distribution shifts with minimal target data. Specifically, we focus on  the extreme case where only single-shot example is available. In such data scarce settings, it is common to leverage synthetic augmentations; examples range from image manipulations to adversarial corruptions~\cite{gokhale2023improving}. Despite their wide-spread adoption, the best augmentation strategy can vary for different shifts, and more importantly, their utility diminishes in the single-shot case. Another popular approach is to use generative augmentations~\cite{SurveyGanAug}, where data variants are synthesized through generative models. Despite being more expressive than generic augmentations, they require comparatively larger datasets for effective training.

\begin{figure*}[t]
    \centering
    \includegraphics[width = 0.99\linewidth]{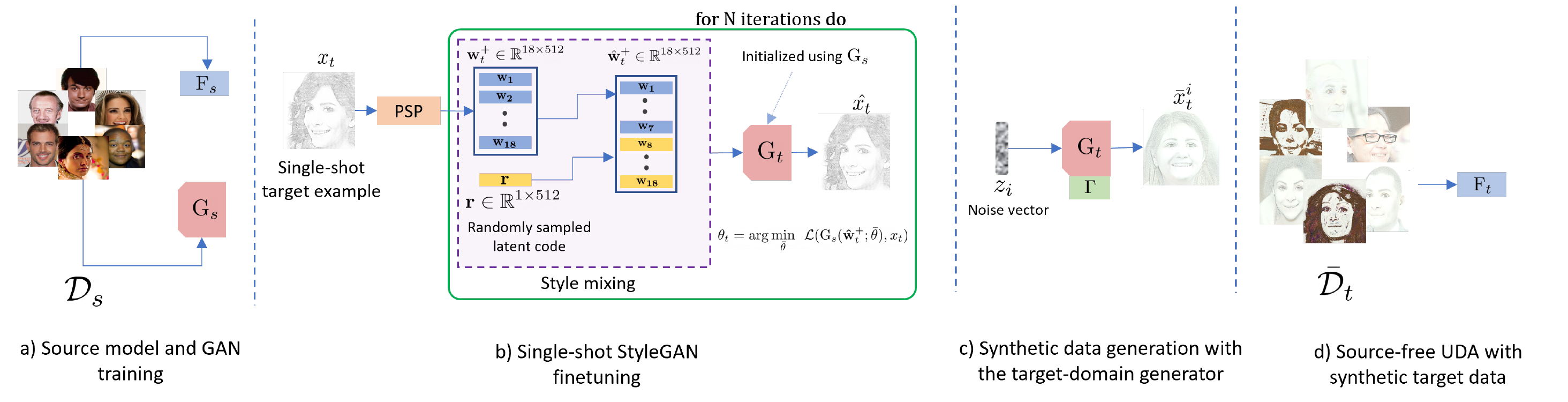}
    \caption{\textbf{\alg}: Assuming access to both the classifier and a StyleGAN from the source domain, we first adapt the generator to the target domain using a single-shot example. Next, we employ the proposed activation pruning strategies to construct the synthetic target dataset $\bar{\mathcal{D}}_t$. Finally, this dataset is used with any SFDA technique for model adaptation.}
    \label{fig:overview}
\end{figure*}
We propose \alg, a new target-aware generative augmentation technique for SFDA with single-shot target data (see Figure \ref{fig:overview}). At its core, \alg~relaxes the assumption of requiring source data, and instead assumes access to a source-trained generative model. We motivate and justify this assumption using a practical vendor-client implementation in Section~\ref{sec:approach}. In this study, we consider StyelGAN as the choice for generative modeling, motivated by their flexibility in disentangling content and style. Our proposed algorithm has two steps, namely \algg~and \algs, to fine-tune a source-trained StyleGAN with the target data, and to synthesize diverse augmentations respectively.



Our contributions can be summarized as follows: 

1. We propose a new target-aware, generative augmentation technique for single-shot adaptation;

2. We introduce two novel sampling strategies based on activation pruning, \textit{prune-zero} and \textit{prune-rewind}, to support domain-invariant feature learning;

3. Using a popular SFDA approach, NRC~\cite{yang2021exploiting}, on augmentations from \alg, we show significant gains in generalization over SoTA online adaptation;

4. By benchmarking on multiple datasets (CelebA, AFHQ, CIFAR-10, DomainNet) and a wide variety of domain shifts (style variations, natural image corruptions), we establish \alg~as a SoTA method for $1-$shot adaptation;

5. We show the efficacy of \alg~in multi-class classification using both class-conditional GANs as well as multiple class-specific GANs.

\section{Background}
\label{sec:background}


\noindent \textbf{Source free domain Adaptation:} In the standard setting of SFDA we only have access to the pre-trained source classifier $\F_s: \x \rightarrow \y$ but not to the source dataset $\mathcal{D}_s=\{(\x_s^i, \y_s^i)\}$. Here, $\x_s^i \in \xs_s$ and $\y_s^i \in \mathcal{Y}$ denote the $i^{\text{th}}$ image and its corresponding label from the source domain $\xs_s$. 
Subsequently, the model needs to be adapted to a target domain $\xs_t$ using unlabeled examples $\mathcal{D}_t = \{(\x_t^j\}$, where  $\x_t^j \in \xs_t$. Note, the set of classes $\mathcal{Y}$ is pre-specified and remains the same across all domains. 


A number of approaches to SFDA have been proposed in the literature and can be categorized into two groups: methods which perform adaptation by fine-tuning the source classifier alone, and those that update the feature extractor as well for promoting domain invariance. In the former category, adaptation is typically achieved through unsupervised/self-supervised learning objectives; examples include rotation prediction~\cite{sun2020test}, self-supervised knowledge distillation~\cite{liu2022source}, contrastive learning~\cite{huang2021model} and batch normalization statistics matching~\cite{Tent,ishii2021source}. The second category includes state-of-the-art approaches such as SHOT~\cite{SHOT}, NRC~\cite{yang2021exploiting} and N2DCX~\cite{tang2021nearest}, which utilize pseudo-labeling based optimization, and often require sufficient amount of data to update the entire feature extractor meaningfully. 

While SHOT is known to be effective under challenging shifts, it relies on global clustering to obtain pseudo-labels for the target data, and in practice, can fail in some cases due to the prediction diversity among samples within a cluster. The more recent NRC~\cite{yang2021exploiting} alleviates this by exploiting the neighborhood structure through the introduction of affinity values that reflect the degree of connectedness between each data point and its neighbors. This inherently encourages prediction consistency between each samples and its most relevant neighbors. Formally, the optimization of NRC involves the following objective:
\begin{equation}
  \mathcal{L}_{\text{NRC}} =   \mathcal{L}_{\text{neigh}}+\mathcal{L}_{\text{self}}+\mathcal{L}_{\text{exp}}+\mathcal{L}_{\text{div}}
\end{equation}where $\mathcal{L}_{\text{neigh}}$ enforces prediction consistency of a sample with respect to its neighbors, while $\mathcal{L}_{\text{self}}$ attempts to reduce the effect of noisy neighbors and $\mathcal{L}_{\text{exp}}$ considers expanded neigbhorhood structure. Finally, $\mathcal{L}_{\text{div}}$ is the widely adopted diversity maximization term implemented as the $KL$ divergence between the distribution of predictions in a batch to a uniform distribution. While \alg~can admit any SFDA technique, we find NRC to be an appropriate choice, since it updates the feature extractor and utilizes the local semantic context to improve performance. This is particularly important in the context of our rich synthetic augmentations, which exhibit a high degree of diversity.



\noindent \textbf{Generative Augmentations:} It is well known that the performance of SFDA methods suffers when the target dataset is sparse. To mitigate this, synthetic augmentations are often leveraged. While it has been found that data augmentation can improve both in-distribution and out-of-distribution (OOD) accuracies~\cite{steiner2021train,hendrycks2021many}, their use in SFDA is more recent. 
Existing augmentations can be broadly viewed in two categories - (i) pixel/geometric corruptions, and (ii) generative augmentations. The former category includes strategies such as CutMix~\cite{cutmix2}, Cutout~\cite{cutout1}, Augmix~\cite{hendrycks2019augmix}, RandConv~\cite{randconv}, mixup~\cite{mixup} and AutoAugment~\cite{autoaugment1}. 
These domain-agnostic methods are known to be insufficient to achieve OOD generalization, especially under complex domain shifts. 
To circumvent this, generative augmentations based on GANs or Variational Autoencoders (VAEs) have emerged. These methods involve training a generative model to synthesize new samples~\cite{SurveyGanAug}. These augmentations have been used in various tasks such as image-to-image translation and improving generalization under shifts. For example, methods such as MBDG~\cite{robey2021model}, CyCADA~\cite{hoffman2018cycada}, 3C-GAN~\cite{Rahman20213CGANCC} and GenToAdapt~\cite{sankaranarayanan2018generate} have leveraged generative augmentations to better adapt to unlabeled target domains. However, by design, these methods require large amounts of data from both source and target domains. In contrast, \alg~focuses on obtaining target-aware generative augmentations by fine-tuning source-trained generative models using only a single-shot target sample. 

\noindent \textbf{StyleGAN-v2 Architecture:} While significant progress has been made in generative AI, including StyleGANs and denoising diffusion models~\cite{saharia2022photorealistic}, we utilize StyleGAN-V2 as the base generative model in our work. This choice is motivated by the flexibility that StyleGANs offer in producing images of different styles, which can be attributed to the inherent disentanglement of style and semantic content in their latent space. Existing approaches works~\cite{wu2021stylespace, wu2021stylealign} have studied this disentanglement property and uncovered the StyleGAN's ability to manipulate the style of an image projected onto the latent space by replacing the latent codes corresponding to only style. Another recent study~\cite{chong2021jojogan} reported that by leveraging such manipulations, one can perform style transfer with a limited number of paired examples. Interestingly, it has also been recently found ~\cite{wu2021stylealign} that, even after transferring a GAN to a different data distribution (faces to cartoons), the latent space of the adapted GAN is point-wise aligned with the source StyleGAN. We take inspiration from these works to develop our single-shot GAN fine-tuning protocol as well as our novel sampling strategies to enable domain-invariant feature learning.
\section{Proposed Approach}
\label{sec:approach}
In this section, we introduce \alg, a new target-aware, generative augmentation strategy with the goal of improving domain adaptation of pre-trained classifiers using single-shot target data. While SFDA methods are known to be effective under a variety of distribution shifts, their performance hinges on the availability of a sufficient amount of target data. In this work, we propose to relax SFDA's assumption on source data access by requiring a source-trained generative model (StyleGANs in our study) to synthesize augmentations in the target domain, in order to enable effective adaptation even under limited data. In particular, we consider the extreme, yet practical setting where only $1-$shot target data is available.

\begin{figure}[t]
    \centering
    \includegraphics[width = 0.99\columnwidth]{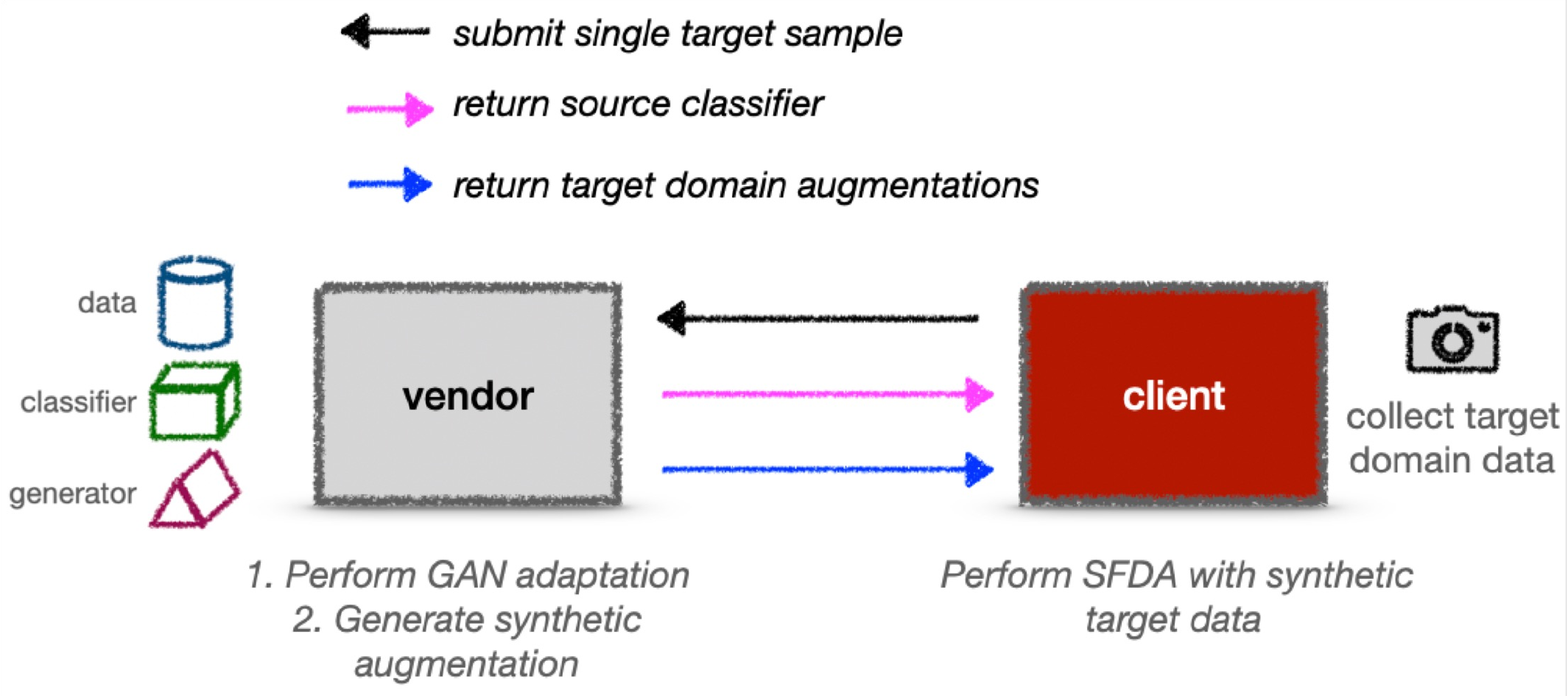}
    \caption{\textbf{A high-level illustration of our adaptation approach} \alg, which is carried out on the \textit{vendor} side that stores the source classifier and a generative model. Designed to support single-shot adaptation, \alg~returns target-aware synthetic augmentations. Finally, the \textit{vendor} executes any SFDA technique to update the source classifier using the synthesized augmentations.}
    \label{fig:implementation}
\end{figure}

Figure \ref{fig:implementation} illustrates an implementation of such a setup where the source dataset, classifier, and the pre-trained generator are available only on the \textit{vendor} side. A \textit{client} that wants to adapt the classifier to a novel domain submits the one-shot target data and receives both the source classifier as well as the synthetic generative augmentations. Finally, the \textit{client} executes any SFDA approach to update the classifier using only the unlabeled synthetic data. This implementation eliminates the need for the \textit{vendor} to share their generative model, while also minimizing the amount of \textit{client} data that gets shared.

As described earlier, \alg~is comprised of two key steps that are carried out on the \textit{vendor} side: (i) \algg: Fine-tune a pre-trained StyleGAN generator $\mathrm{G}_s$ using single-shot target data $\{\x_t\}$ under unknown distribution shifts.; and (ii) \algs: Synthesize diverse samples $\mathcal{D}_t = \{\bar{\x}_t^j\}$ using the fine-tuned generator $\mathrm{G}_t$ to support effective classifier adaptation to the target domain. Finally, we leverage the recently proposed NRC method to perform \textit{client}-side adaptation. Now, we describe these steps in detail.

\begin{algorithm}[t]
\caption{\algg}
\label{algo1}
\begin{algorithmic}[1]
\STATE \textbf{Input}:~Target sample $\x_t$, No. of training iterations $M$,\\ \quad \quad \quad
	        Source generator $\mathrm{G}_s$, Inversion module $\mathcal{E}$ \\ \quad \quad \quad Set of style layers $\mathcal{L}_{\text{st}}$.
         
\STATE \textbf{Output}: Fine-tuned generator $\mathrm{G}_t$.
\STATE Invert the target sample to obtain  $\mathbf{w}_t^+ = \mathcal{E}(\x_t)$\;
    \FOR{$m$ in $1$ to $M$}
        \STATE Generate random style latent $\mathbf{r}^+$\;
        \STATE Perform style-mixing, \textit{i.e.}, replace style layers $\mathcal{L}_{\text{st}}$ of $\mathbf{w}^+_t$ with $\mathbf{r}^+$\; \\
        \STATE Generate image $\hat{\x}_t = \mathrm{G}_s(\mathbf{\hat{w}}^+_t)$\; \\
        \STATE Update parameters $\Theta_t$ using \eqref{eq:ganInversion}\;
    \ENDFOR
       
    \STATE \textbf{return}:~$\mathrm{G}_t$ with parameters $\Theta_t$.
\end{algorithmic}

\end{algorithm}

\subsection{\algg: Single-Shot StyleGAN Fine-Tuning} Our goal in this step is to fine-tune $\mathrm{G}_s$ using only the single-shot example $\x_t$ from the target domain to produce an updated generator $\mathrm{G}_t$. To this end, the proposed approach first inverts $\x_t$ onto the style-space of $\mathrm{G}_s$. In practice, this can be done using one of the following strategies: (i) a pre-trained encoder such as Pixel2Style2Pixel~\cite{richardson2021encoding} or E4E~\cite{tov2021designing}, which maps a given image into the style code $\mathbf{w}_t^+ \in \mathbb{R}^{L\times512}$. This latent code corresponds to $L$ intermediate layers of a StyleGAN model (e.g., $L = 18$ in StyleGAN-v2); (ii) any standard GAN inversion technique to infer an approximate solution in the style space~\cite{xia2022gan}; (iii) text-guided inversion such as StyleClip~\cite{patashnik2021styleclip} if the label is available for the single-shot target image. Though conventional GAN inversion is known to be expensive, it will not be a significant bottleneck with only a single image.

Without loss of generality, the target domain is expected to contain distribution shifts w.r.t. the source domain, and hence the inverted solution in the style-space is more likely to resemble the source domain. For example, inverting a cartoon into the style-space of a GAN trained on real face images will produce a semantically similar image from the face manifold. Recent evidence~\cite{subramanyam2022improved} suggests that one can accurately recover an OOD image using an additional vicinal regularization to the inversion process. However, in our case, we do not want an accurate reconstruction, but rather refine the generator $\mathrm{G}_s$ to emulate the characteristics of a target domain.

To this end, we utilize the following loss function defined on the activations from the source-domain discriminator $\mathrm{H}_s$: 
\begin{align}
   \Theta_t = \arg \min_{\bar{\Theta}} \phantom{r} \sum_{\ell} \|\mathrm{H}^{\ell}_s(\mathrm{G}_s(\mathbf{w}_t^+;\bar{\Theta})) - \mathrm{H}_s^{\ell}(\x_t)\|_1,
    \label{eq:ganInversion}
\end{align}where $\mathbf{w}_t^+$ is the style-space latent code obtained via GAN inversion, $\Theta_t$ refers to the parameters of the updated generator $\mathrm{G}_t$ and $\mathrm{H}^{\ell}_s$ denotes the activations from layer $\ell$ of the discriminator $\mathrm{H}_s$. Intuitively, this objective minimizes the discrepancy between the target image and the reconstruction from the updated generator. Note that, the parameters of the discriminator are not updated during this optimization. While any pre-trained feature extractor can be used for this optimization, the source discriminator provides meaningful gradients by comparing both the content and style aspects of the target image. Upon training, we expect the generator $\mathrm{G}_t$ to produce images resembling the target domain for any random latent code in the style-space.

An inherent issue with our objective is that, this optimization can be highly unstable when using a single $\x_t$. To circumvent this, we leverage multiple, style-manipulated versions of $\x_t$ through a style-mixing protocol. More specifically, we first generate a random code $\mathbf{r}^+$ in the style-space (using the mapping network in StyleGAN). Next, we perform mixing by replacing the latent codes from a pre-specified subset of layers $\mathcal{L}_{\text{st}}$ in $\mathbf{w}_t^+$ using the corresponding codes from $\mathbf{r}^+$. In effect, this produces a modified image that contains the content from $\mathbf{w}_t^+$ and the style from $\mathbf{r}^+$. We denote this style-manipulated latent using the notation $\mathbf{\hat{w}}_t^+$. In each iteration of our optimization, a different style-mixed latent code $\mathbf{\hat{w}}_t^+$ is generated  to compute the loss in \eqref{eq:ganInversion}. Algorithm \ref{algo1} summarizes the steps of \algg.

\noindent \textbf{Choosing layers for style-mixing.} We choose $\mathcal{L}_{\text{st}}$ by exploiting the inherent style and content disentanglement in StyleGANs. Priors works~\cite{wu2021stylespace, kafri2021stylefusion, karras2020analyzing} have established that the initial layers typically encode the semantic content, while the later layers capture the style characteristics. Since the exact subset of layers that correspond to style vary as the image resolution changes, following standard practice, we used $\mathcal{L}_{\text{st}} = {8-18}$ when $\mathrm{G}_s$ produces images of size $1024 \times 1024$ and $\mathcal{L}_{\text{st}} = {3-8}$ for images of size $32 \times 32$ (CIFAR-10).

\subsection{\algs: Target-aware Augmentation Synthesis}
Once we obtain the target domain-adapted StyleGAN generator $\mathrm{G}_t$, we next synthesize augmentations by sampling in its latent space. Despite the efficacy of such an approach, the inherent discrepancy between the true target distribution $P_t(\x)$ and the approximate $Q_t(\x)$ (synthetic data) can limit generalization. Existing works~\cite{kundu2020towards} have found that constructing generic representations (using standard augmentations) is useful for test-time adaptation any domain. However, in contrast, our goal is to produce augmentations specific only to a given target domain, thus enabling effective generalization even with single-shot data. 

\begin{algorithm}[t]
\caption{\algs}
\label{algo2}
\begin{algorithmic}[1]
\STATE \textbf{Input}:~Target GAN $\mathrm{G}_t(.;\Theta_t)$, Source GAN $\mathrm{G}_s(.;\Theta_s)$, \\ \quad \quad \quad 
Pruning strategy $\mathrm{\Gamma}$,	Pruning ratio $p$,\\ \quad \quad \quad Set of style layers $\mathcal{L}_{\text{st}}$;
\STATE \textbf{Output}:~Sampled image $\bar{\x}_t$
    \STATE Draw a random latent code  $\mathbf{w}^+$ from $\mathrm{G}_t(.;\Theta_t)$\;
    \FOR{$\ell$ in $\mathcal{L}_{\text{st}}$}
        \STATE $\beta \sim \text{RandInt}(0,1)$\;
        \IF{$\beta==1$}
            \STATE Obtain layer $\ell$ activations $\mathrm{h}_t^{\ell}$ from $\mathrm{G}_t(\mathbf{w}^+)$\;
            \STATE \textcolor{gray}{/* Iterate over activation channels $V^{\ell}$*/}
            \FOR{$v$ in $1$ to $V^{\ell}$}
                \STATE $\tau_p = p$-th percentile of $\mathrm{h}_t^{\ell}[:,:,v]$\;
                \IF{$\mathrm{\Gamma}==$ \textcolor{blue}{\textit{prune-zero}}}
                    \STATE $\mathrm{h}_t^{\ell}[i,j,v] = 0$ \textbf{if} $\mathrm{h}_t^{\ell}[i,j,v] < \tau_p, \forall~i, j$\;
                \ELSE
                    \STATE Obtain activations $\mathrm{h}_s^{\ell}$ from $\mathrm{G}_s(\mathbf{w}^+)$\;  \\                 
                     \STATE \resizebox{.7 \columnwidth}{!} 
                        {
                           $\mathrm{h}_t^{\ell}[i,j,v] = \mathrm{h}_s^{\ell}[i,j,v]$ \textbf{if} $\mathrm{h}_t^{\ell}[i,j,v] < \tau_p, \forall~i, j$
                        }\; \\
                \ENDIF
            \ENDFOR
        \ENDIF 
    \ENDFOR
       
    \STATE \textbf{return}:~Image $\bar{\x}_t = \mathrm{G}_t(\mathbf{w}^+; \mathrm{\Gamma})$

\end{algorithmic}

\end{algorithm}

To this end, we propose two novel strategies that perturb the latent representations from different layers of $\mathrm{G}_t$ to realize a more diverse set of style variations. Both our sampling strategies are based on activation pruning, \textit{i.e.}, identifying the activations in each style layer that are lower than the $p^\text{th}$ percentile value of that layer, and replacing them with (i) zero (referred to as \textit{prune-zero}); or (ii) activations from the corresponding layer of the source GAN $\mathrm{G}_s$ (\textit{prune-rewind}). The former strategy aims at creating a generic representation by systematically eliminating style information in the image. On the other hand, the latter attempts to create a smooth interpolation between the source and target domains by mixing the activations from the two generators. Note, we perform pruning only in the style layers, so that the semantic content of a sample is not changed. Note, we use the same set of style layers selected for performing \algg. Algorithm \ref{algo2} lists the activation pruning step.

\subsection{\alggmc: Extending to class-conditional GANs} 
When dealing with multi-class problems, it is typical to construct class-conditional GANs, $\mathrm{G}_s(.;\text{c})$, to effectively model the different marginal distributions. In such settings, images from different classes get mapped to disparate sub-manifolds in the StyleGAN latent space. Assuming there are $K$ different classes in $\ys$, we can directly apply \algg~using 1-shot examples from each of the classes. The only difference occurs in the GAN inversion step, wherein we need to identify the conditioning variable $\text{c}$ along with the latent code $\mathbf{w}_t^+$. Note, if the labels are available, one can estimate only $\mathbf{w}_t^+$. Finally, the algorithm \ref{algo1} is repeated with $K$ target images. We refer to this protocol as \alggmc~(multi-class generation).

However, when we perform \alggmc~using only a subset of the classes (say only one out $K$), there is a risk of not incorporating target-domain characteristics into the images synthesized for all realizations from the latent space. However, as we will show in the results (Figure \ref{fig:cifar}), even using an example from a single class still leads to significantly improved generalization. We hypothesize that this behavior is due to the fact that the synthesized augmentations (random samples from $\mathrm{G}_t$) arise from both $\xs_s$ and $\xs_t$, thus emulating an implicit mixing between the two data manifolds.

    

\begin{figure}[t]
    \centering
    \includegraphics[width=0.99\columnwidth]{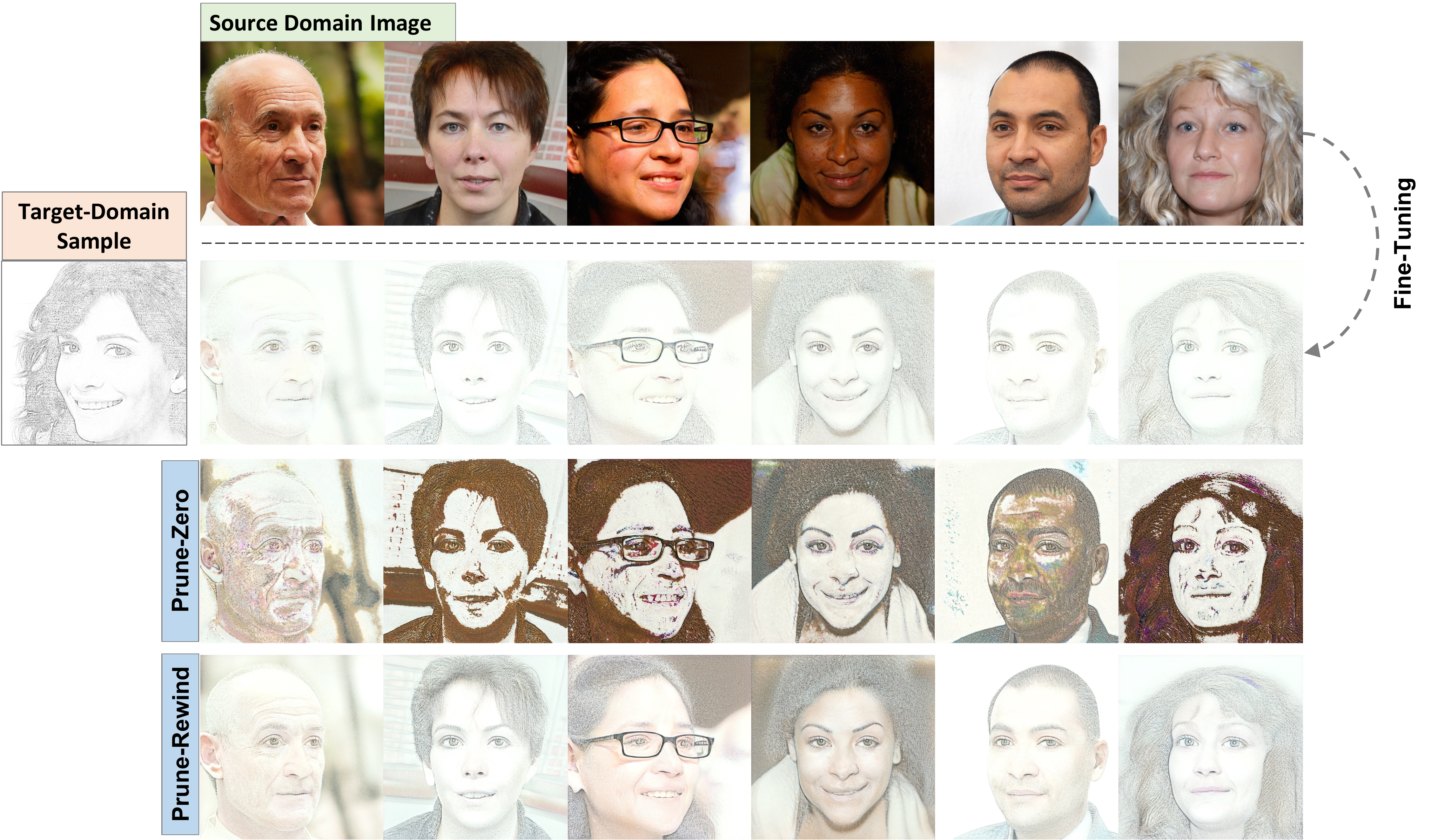}
    \caption{\textbf{Synthetic data generated using our proposed approach}. In each case, we show the source domain image and the corresponding reconstructions from the target StyleGAN sampling (base), prune-zero and prune-rewind strategies.}
    \label{fig:augs}
    \vspace{-0.1in}
\end{figure}

\section{Experiments}
\label{sec:results}
 We perform an extensive evaluation of \alg~using a suite of classification tasks with multiple benchmark datasets, different StyleGAN architectures and more importantly, a variety of challenging distribution shifts. In all our experiments, we use single-shot target data and utilize publicly available, pre-trained StyleGAN weights.

\subsection{Experimental Setup} 

\begin{figure*}[t]
    \centering
    \includegraphics[width=0.75\linewidth]{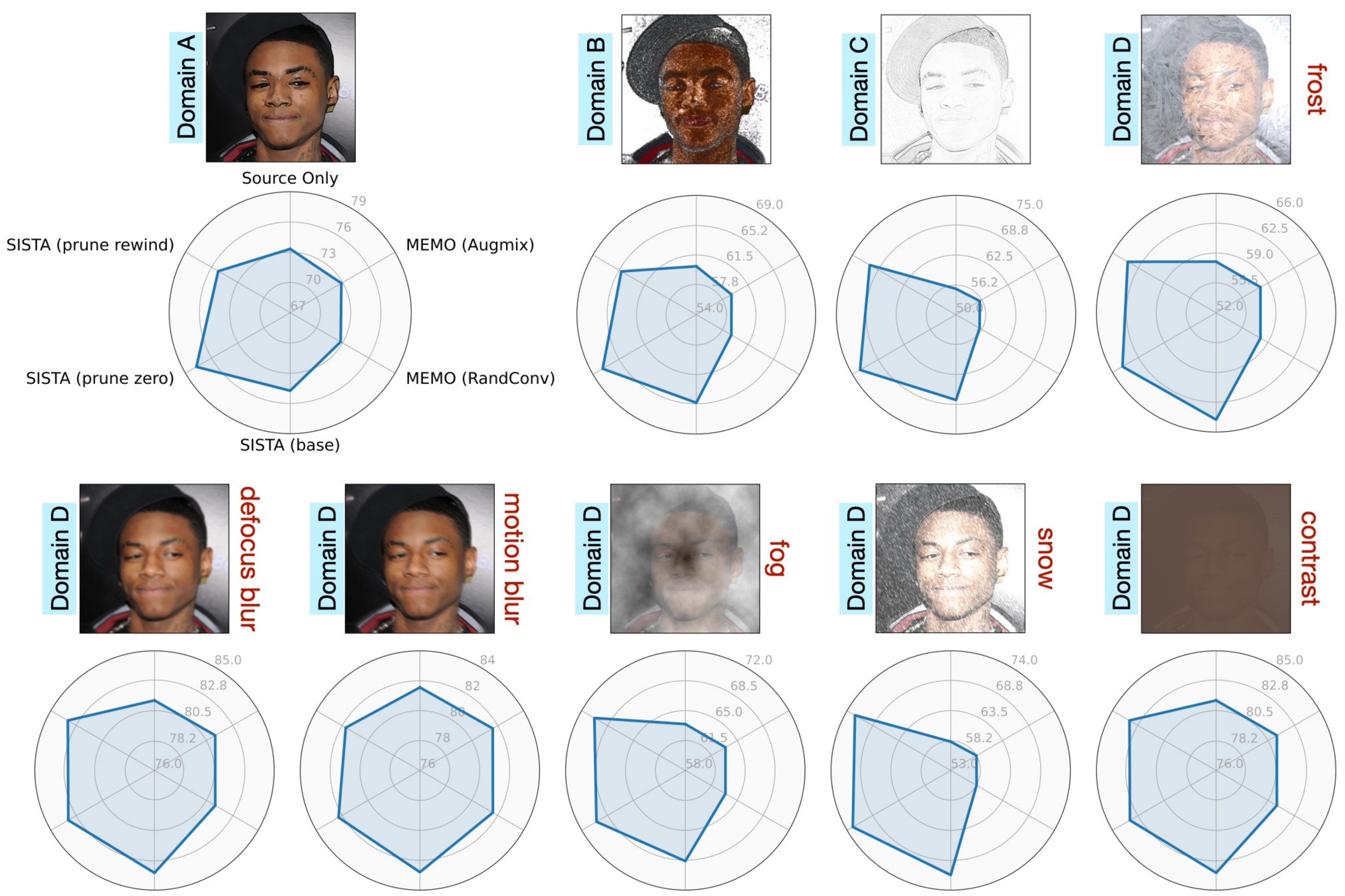}
    \caption{\textbf{\alg~significantly improves generalization of face attribute detectors}. We report the $1-$shot SFDA performance (Accuracy \%) averaged across different face attribute detection tasks for different distribution shifts (Domains A, B \& C) and a suite of image corruptions (Domain D). \alg~consistently improves upon the baseline(source-only) and SoTA baseline MEMO in all cases. }
    \label{fig:spiders}
\end{figure*}
    
\noindent{\textbf{Datasets:}} For our empirical study, we consider the following four datasets: (i) CelebA-HQ~\cite{DBLP:journals/corr/abs-1710-10196} is a high-quality (1024x1024 resolution) large-scale face attribute dataset with $30$K images. We split this into a source dataset of $18$K images and the remaining was used to design the target domains. We perform attribute detection experiments on a subset of $19$ attributes, i.e., each attribute is posed as its own binary classification task; (ii) AFHQ~\cite{choi2020starganv2} is a dataset of animal faces consisting of 15,000 images at 512×512 resolutions with three classes, namely cat, dog and wildlife, each containing 5000 images. For each class, $500$ images were used to create the target domains, and the remaining was used as the source data; (iii) CIFAR-10~\cite{krizhevsky2009learning} is also a multiclass classification dataset with 60000 images at 32x32 resolution from 10 different object classes. We use the standard train-test splits for constructing the source and target domain datasets. While we used the StyleGAN-v2 trained on FFHQ faces for our experiments on the CelebA-HQ dataset\footnote{\tiny{\url{https://github.com/rosinality/stylegan2-pytorch}}}, for AFHQ and CIFAR-10 we obtained the pre-trained StyleGAN2-ADA models\footnote{\tiny{\url{https://github.com/NVlabs/stylegan2-ada-pytorch }}} from their respective sources; and (iv) DomainNet~\cite{domainnet}, a large-scale benchmark comprising $6$ domains namely Clipart, Painting, Quickdraw, Sketch, Infograph and Real with each domain consisting of images from $340$ categories. For this experiment, we used the state-of-the-art StyleGAN-XL model~\cite{sauer2022stylegan} trained on ImageNet~\cite{ILSVRC15}. Note, we used only the subset of categories from DomainNet that directly overlapped with ImageNet classes. To the best of our knowledge, this is the first work to report adaptation performance with a single target image on DomainNet, and to use ImageNet-scale StyleGAN-XL for data augmentation.

\noindent{\textbf{Target Domain Design:}} To emulate a wide-variety of real-world shifts, we employed standard image manipulation techniques (we will release this new benchmark dataset along with our codes) to construct the following target domains: (i) \textit{Domain A}: We used the \textit{Stylization} technique in OpenCV with $\sigma_s=40$ and $\sigma_r = 0.2$; (ii) \textit{Domain B}: For this shift, we used the \textit{PencilSketch} technique in OpenCV with $\sigma_s=40$ and $\sigma_r = 0.04$;  (iii) \textit{Domain C}: This challenging domain shift was created by converting each color image to grayscale, and then performing pixel-wise division with a smoothed, inverted grayscale image; and (iv) \textit{Domain D}: This shift was created using a different natural image corruptions from ImageNet-C~\cite{hendrycks2019robustness} typically used for evaluating model robustness. In particular, we used the \textit{imagecorruptions}\footnote{\tiny{\url{https://github.com/bethgelab/imagecorruptions }}} package for realizing $6$ different shifts, namely \textit{contrast, defocus blur, motion blur, fog, frost and snow}. We report our performance across all the domain shifts for the different attribute detection tasks. Given the inherently challenging nature of Domain C, we used that exclusively to evaluate the multi-class classifiers trained on AFHQ and CIFAR-10 datasets. Finally, for DomainNet evaluations we considered \textit{Real photos} as the source domain and used each of the five remaining domains as the target.

    \begin{figure*}[t]
    \centering
    \subfloat[c][CIFAR-10.]{\includegraphics[width=0.6\linewidth]{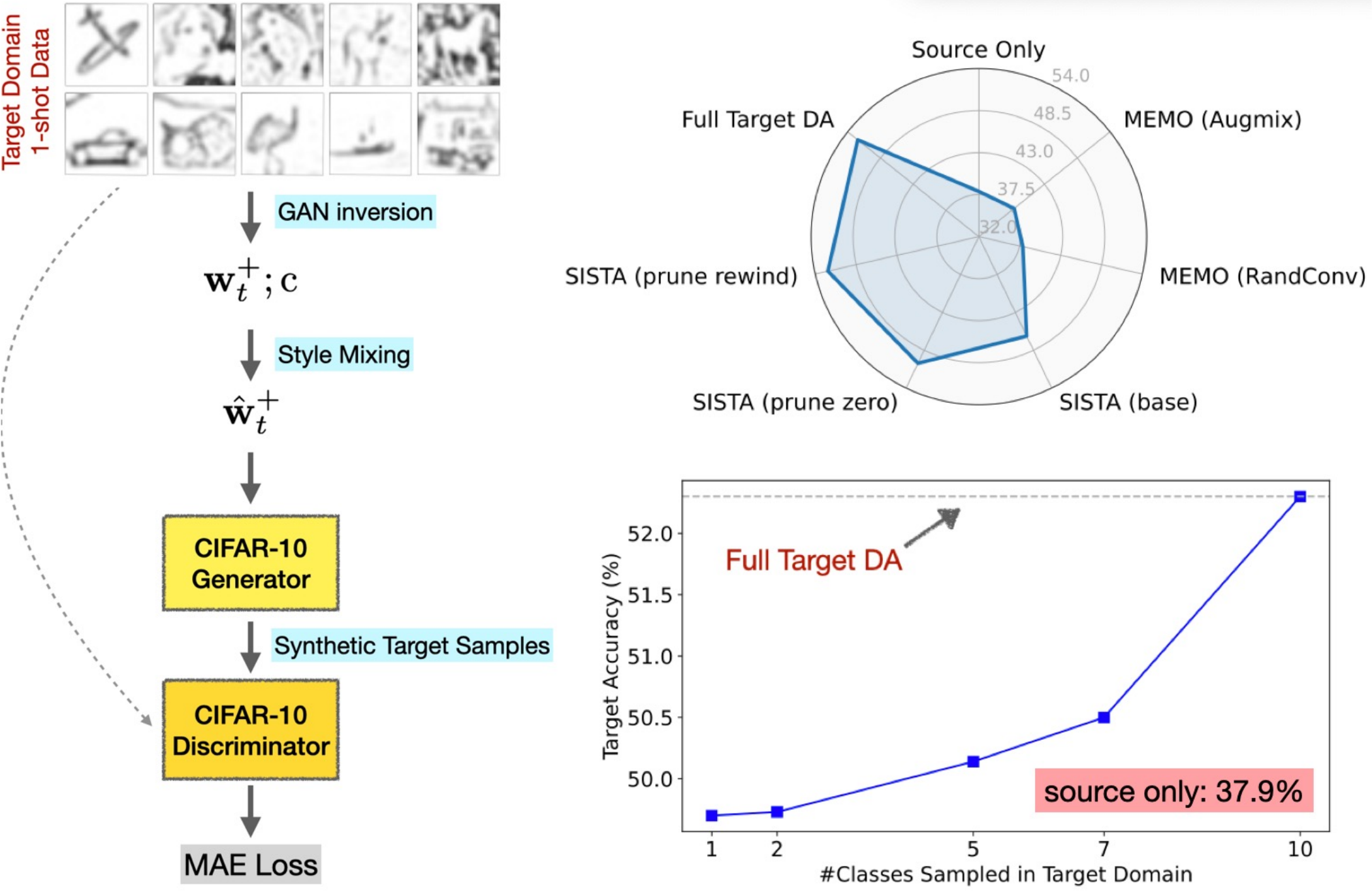}
     \label{fig:cifar}}
     \hfill
     \subfloat[c][AFHQ.]{\includegraphics[width=0.31\linewidth]{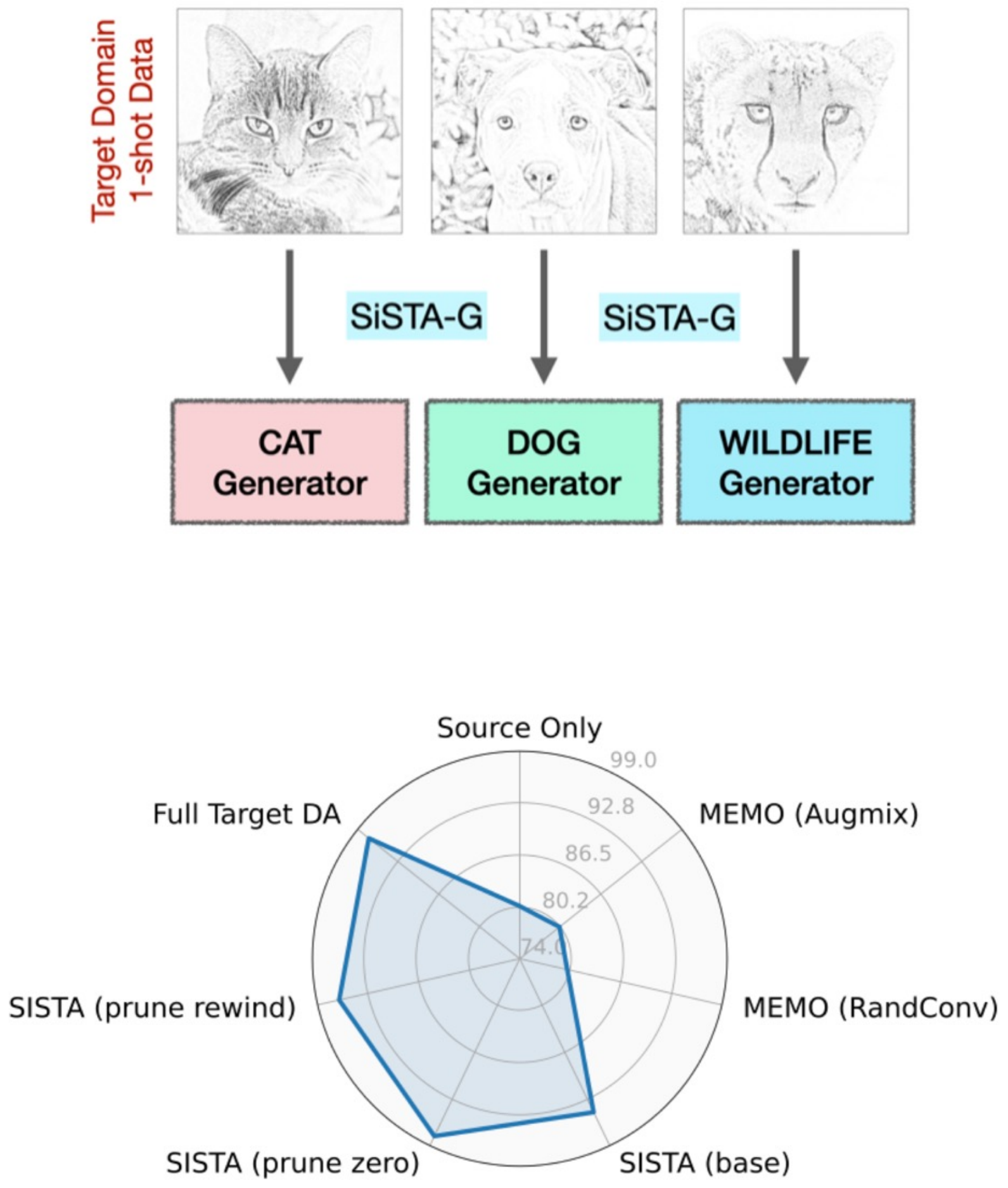}
     \label{fig:afhq}}
    \caption{\textbf{Multi-class classification}: (a)-left illustrates \alggmc~with class-conditoned GANs, (a)-right shows the performance of~\alg, while the bottom plot studies the performance of\alg~with exposure to only a subset of  classes from the target domain. (b) visualizes our approach for AFHQ dataset where individual class-specific generators are finetuned and bottom plot analyses \alg~along with baselines for this challenging dataset. }
    \label{fig:multi}

\end{figure*}
    
\noindent{\textbf{Evaluation methodology:}}
        (a) \underline{\textit{Source model training}}: To obtain the source model $\mathrm{F}_s$ we fine-tune an ImageNet pre-trained ResNet-$50$~\cite{he2016deep} with labeled source data. We use a learning rate of $1e-4$, Adam optimizer and train for $30$ epochs;        
        (b) \underline{\textit{StyleGAN fine-tuning}}: We fine-tune $\mathrm{G}_s$ for $300$ iterations ($M$ in Algorithm \ref{algo1}) using one-target image with learning rate set to $2e-3$ and Adam optimizer with $\beta=0.99$.  These parameters were identified using the CelebA benchmark and we used the same settings for all experiments;
        (c) \underline{\textit{Synthetic data curation}}: The size of the synthetic target dataset $\bar{\mathcal{D}}_t$, $T$, was set to $1000$ images in all experiments. Note, in section \ref{sec:analysis}, we study the impact of this choice. 
        Another important hyperparameter is the choice of GAN layers for style manipulation: (i) layers $8-18$
 in StyleGAN-2; (ii) layers $3-8$
 in CIFAR-10 GAN; (iii) layers $10-27$
 in StyleGAN-XL. This selection was motivated by findings from recent studies on style/content disentanglement in StyleGAN latent spaces~\cite{wu2021stylespace,kafri2021stylefusion, karras2019style}.
        (d) \underline{\textit{Choice of pruning ratio}}: For all experiments, we set $p = 20\%$ for prune-rewind and $p = 50\%$ for prune-zero strategies. Note, in section \ref{sec:analysis}, we study the impact of this choice;
        (e) \underline{\textit{SFDA training}}: For the NRC algorithm, we set both neighborhood and expanded neighborhood sizes at $5$ respectively. Finally, we adapt $\mathrm{F}_s$ using SGD with momentum $0.9$ and learning rate $1e-3$. All results that we report are computed as an average of $3$ independent trials;
        (f) For evaluation, we report the target accuracy (\%) on a held-out test set in each of the target domains.
        
\noindent{\textbf{Baselines:}} In addition to the vanilla source-only baseline (no adaptation), while there exists a number of test-time adaptation approaches, we perform comparisons to the state-of-the-art online adaptation method, MEMO~\cite{zhang2021memo}, that enforces prediction consistency between an image and its augmented variants. In particular, we implement MEMO with two popular augmentation strategies namely Augmix and RandConv~\cite{randconv}. We choose MEMO as the key baseline, since it is already well established that it is superior to other protocols like TENT and TTT. Finally, for comparison, we report the Full Target DA performance as an upper bound, \textit{i.e.}, when the entire target dataset (unlabeled) is used for adaptation.

\subsection{Findings}
Figure \ref{fig:augs} illustrates the synthetic data generated for a target domain (\textit{pencil sketch}) using vanilla sampling (or base), \textit{prune-zero} and (\textit{prune-rewind}) strategies. More examples can be found in the supplement (Figure \ref{fig:domain_augs}).

\noindent{\textbf{\alg~consistently produces superior performance across different distribution shifts.}}

In Tables~\ref{tab:DomainA}-\ref{tab:contrast}, the performance of \alg~across different domain shifts (A, B, C, D) on the CelebA-HQ dataset is compared to the baselines for all the $19$ attributes. Furthermore, Figure \ref{fig:spiders} summarizes the average performance (across attributes and multiple trials) for the CelebA-HQ dataset. We see that when compared to the source-only baseline and the state-of-the-art MEMO, \alg~yields average improvements of $4.41\%$, $7.5\%$, $17.73\%$ and $5.1\%$ respectively for the four target domains. This improvement can be directly attributed to the efficacy of our proposed augmentations, which enable the SFDA method to learn domain-invariant features when adapting the source classifier.

Additionally, utilizing the proposed activation pruning strategies reveal significant gains under severe shifts over the na\"ive sampling (base). For example, we see an average improvement of $18\%$ across different attributes in Domain C, when compared to the state-of-the-art MEMO. In particular, we notice that for challenging attributes such as \textit{bangs}, \textit{blond hair}, and \textit{gender}, we obtain striking $26.1\%, 29.6\%, 33.9\%$ improvements over the source-only performance. This illustrates how our pruning strategy can create generic representations that aid in an effective adaptation.

\noindent\textit{\textbf{Failure cases}}: While \alg~is generally very effective, there are a few cases where it does not perform as expected. For example, with the Domain B results in Table~\ref{tab:DomainB}, we notice that for certain attributes (\textit{$5$'o clock shadow}, \textit{bald}), we fail to improve over the source-only performance (near-random performance), since it becomes challenging to resolve those attributes under that distribution shift. Additionally, in Domain C, we find that the performance of \alg~(base) is sometimes greater than that of \alg~(prune zero), likely due to the excessive elimination of style information during pruning. While this can be potentially fixed by adjusting the prune ratio or increasing the number of augmented samples (see \ref{sec:analysis}), this reveals some of the failure scenarios for \alg.


\noindent{\textbf{\alg~can handle natural image corruption.}} Natural image corruptions mimic domain shifts that are prevalent in real-world settings. Surprisingly, we find that our proposed \algs~protocol is able to fine-tune the GAN even under such image corruptions and lead to apparent gains in the generalization performance. More specifically, we want the emphasize the two challenging corruptions, namely contrast and fog, where the class discriminative features appear to be muted. Even under these corruptions, as showed in Figure \ref{fig:spiders}, \alg~achieve average performance improvements of $10.14\%$ and $6.52\%$, respectively. 

\begin{table}[t]
\centering
\caption{Performance of \alg~on the five different domains of the DomainNet Dataset. \alg~consistently improves over the Source Only and MEMO baselines even under such complex domain shifts.}
\renewcommand{\arraystretch}{1.5}
\resizebox{\linewidth}{!}{
\begin{tabular}{|l|c|c|c|c|c|}
\cline{2-6}
\multicolumn{1}{c|}{}                                 & QuickDraw & Painting & ClipArt & InfoGraph & Sketch\\
\cline{2-6} \hline
\cellcolor[HTML]{FFE6E6} Source only           & 9.23      & 62.25    & 58.55   & 28.45     & 43.86  \\\hline \hline
MEMO (Augmix)        & 8.73      & 62.20    & 60.15   & 28.61     & 43.86  \\\hline
MEMO (RandConv)      & 8.04      & 61.91    & 59.23   & 28.02     & 43.52  \\\hline \hline
SiSTA ( base)        & 11.78     & 63.53    & 60.98   & 31.61     & 47.54  \\\hline 
SiSTA (prune-zero)   & \textbf{13.12}     & 63.69    & 60.98   & 31.65     & \textbf{48.12}  \\\hline 
SiSTA (prune-rewind) & 11.86     & \textbf{64.05}    & \textbf{61.02}   & \textbf{31.8 }     & 46.78  \\\hline \hline
\cellcolor[HTML]{E6FFF2} Full Target DA       & 16.27     & 68.99    & 69.55   & 31.77     & 55.09 \\\hline
\end{tabular}
}
\label{tab:domainnet}
\vspace{-0.1in}
\end{table}

 \noindent{\textbf{\alg~is effective even with class-conditional GANs.}} 
 In this experiment, we study how \alg~performs on CIFAR-10 adaptation, when we are provided with a class-conditional StyleGAN. In this case, we use the \alggmc~procedure to perform GAN fine-tuning, which requires the GAN inversion step to identify both the latent code as well as the conditioning variable. As illustrated in Figure \ref{fig:cifar}, we use $1-$shot examples from each of the $10$ classes and synthesize $T=1000$ augmentations from \alg. Note, during sampling, we draw from the different classes randomly. We find that, for the challenging Domain C target, \alg~not only outperforms the baselines by a large margin, but also matches the Full Target DA performance, while using only a single-shot example. Furthermore, as argued in Section 3.3, using single-shot examples from even a subset of classes can be beneficial. To demonstrate this, we varied the number of classes from which target examples are drawn ($1$ to $10$). We find that, even with a single class example, \alg~provides a large gain of $12.69\%$ over the source-only baseline. As expected, the generalization performance consistently improves as we expose the model to examples from additional classes.
 \begin{figure}[t]
\centering
\subfloat[c][Varying prune ratio]{\includegraphics[width=0.99\columnwidth]{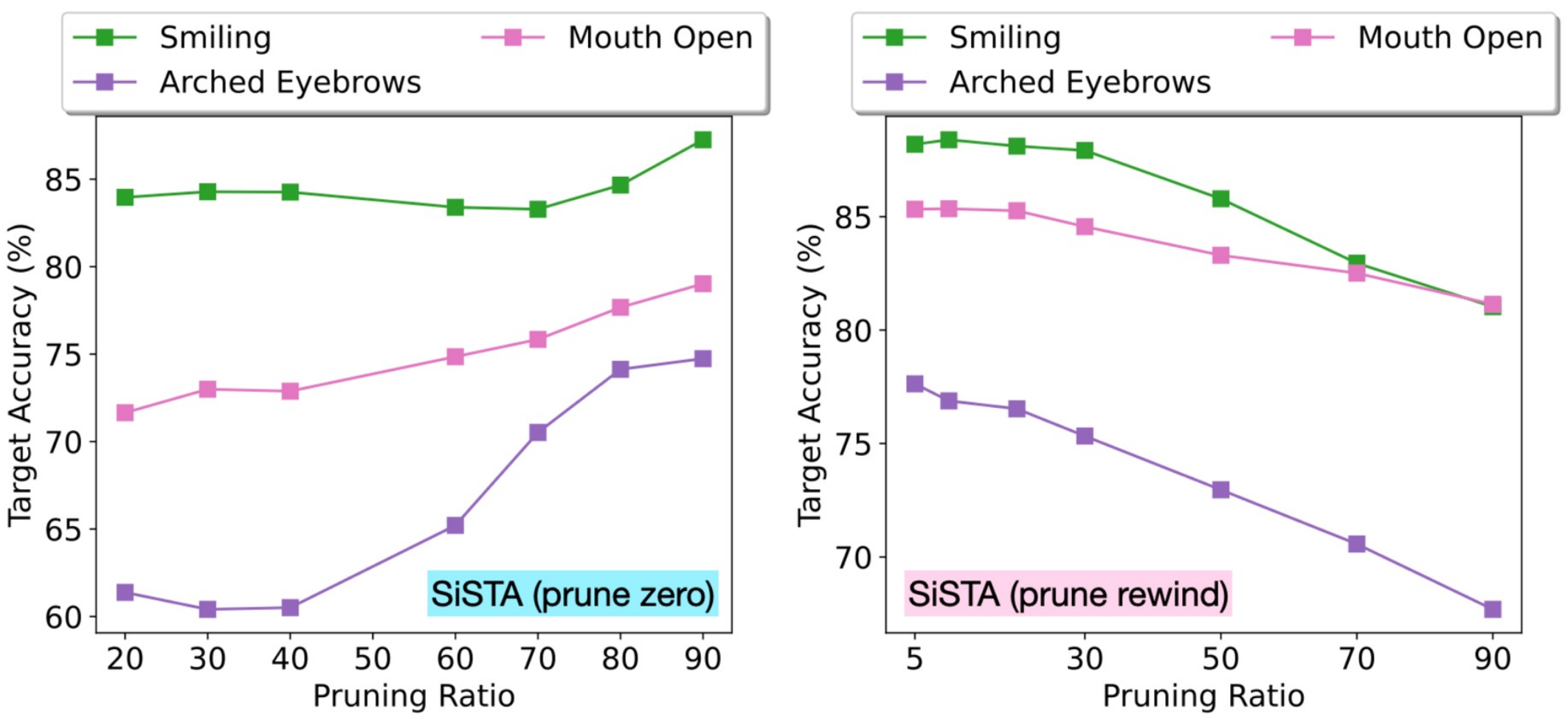}
 \label{fig:prune}}
 \vfill
 \subfloat[c][Varying T]{\includegraphics[width=0.99\columnwidth]{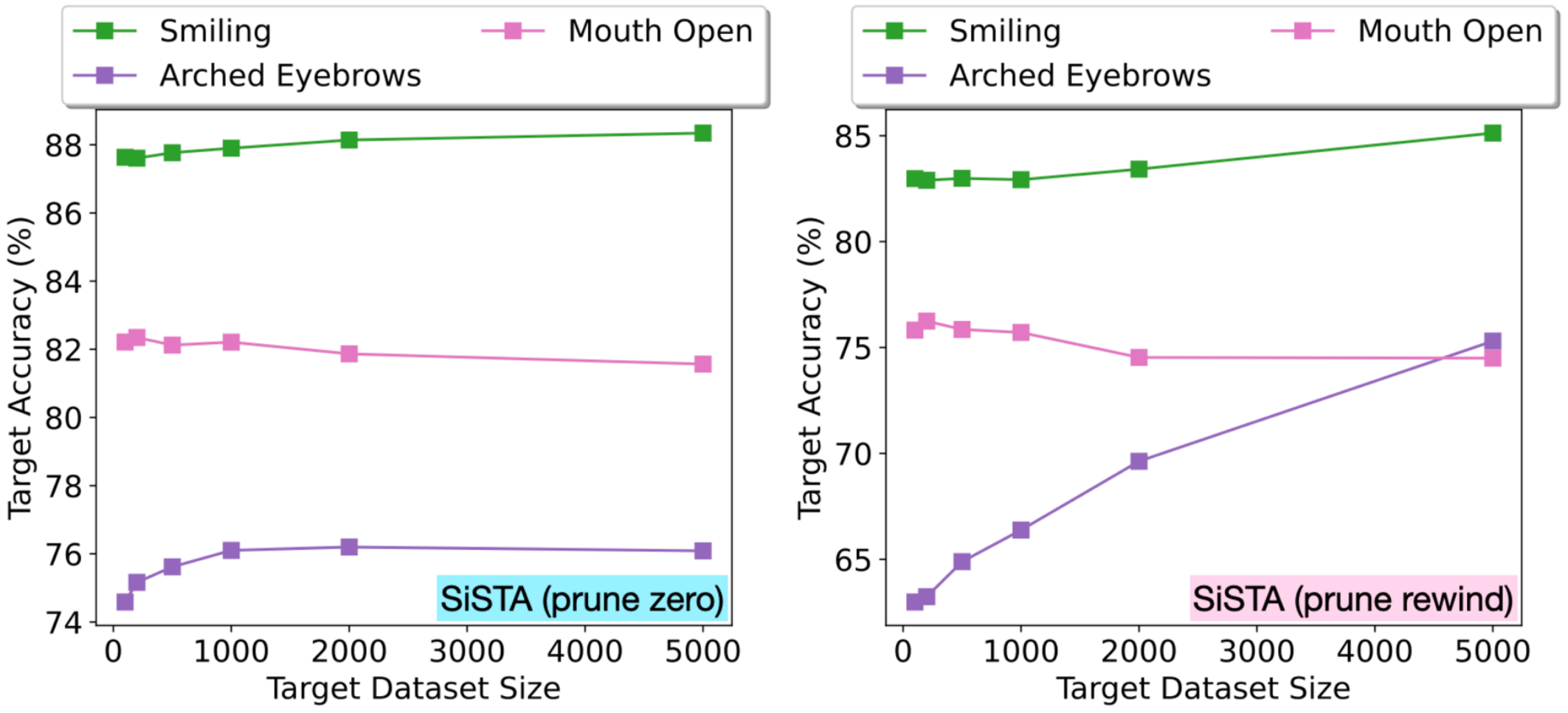}
 \label{fig:varyT}}
\caption{\textbf{Analysis} of varying prune ratio $p$ and the amount of synthetic target domain data $T$ used by \alg.}
\vspace{-0.2in}
\end{figure}

 \noindent{\textbf{\alg~can also be used with multiple class-specific GANs.}} In this study, we examined the performance of \alg~in a multi-class classification problem with AFHQ, where we assume access to individual generative models for each class. Given the inherent diversity within classes (different breeds of cats or dogs), it is sometimes challenging to train a single StyleGAN for the entire data distribution. In such cases, a separate generative model can be trained on source images from each of the classes. However, the classifier is trained for a $3-$way classification setting. In this case, we perform \alg~for each GAN independently using its corresponding example. As shown in Figure \ref{fig:afhq}, we find that, even our base variant achieves $94.53\%$, outperforming the source-only and baselines by large margins ($14\%$). Our best performance is achieved by \textit{prune-zero} in this setting and it matches Full Target DA.

 \begin{figure*}[t]
    \centering
    \includegraphics[width=\linewidth]{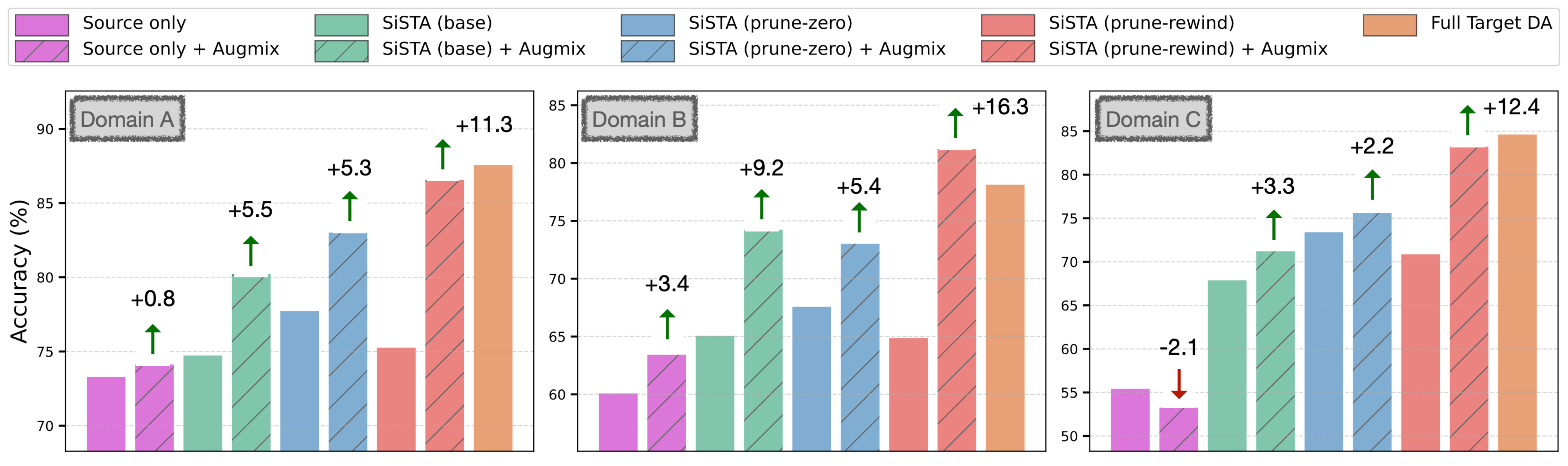}
\caption{\textbf{Effect of Toolbox augmentations on \alg.} We present the performance of \alg~on Domains A, B, and C of the CelebaHQ dataset when images generated by \alg~are further enhanced with Augmix~\cite{hendrycks2019augmix}. We observe that toolbox augmentations can further improve the performance of \alg, and in a few cases, \alg~even surpasses the Full Target DA baseline.}
    \label{fig:sista_augmix}
    \vspace{-0.15in}
\end{figure*}
 
\noindent{\textbf{Even on large scale benchmarks such as DomainNet, \alg~provides consistent benefits.}} To study its performance on large-scale benchmarks, we tested \alg~on DomainNet that comprises a large number of object types and complex distribution shifts (photo, quickdraw, painting, etc.). Given the diversity of objects in this benchmark, we utilized the state-of-the-art StyleGAN-XL model trained on ImageNet to perform \alg~and studied the single-shot adaptation performance for different target domains (real is the source domain). From Table \ref{tab:domainnet}, we find that even on this benchmark, \alg~(prune-zero) convincingly improves upon source only baselines. For example, \alg~provides about 4\% improvements for Quickdraw and Sketch domains. As with the other benchmarks, \alg~is indeed competitive to the Full Target DA baseline. 

\subsection {Analysis of parameter choices}
\label{sec:analysis}
\textbf{\textbf{The choice of prune ratio $p$.}}  We investigate the effect of the choice for $p$ in \textit{prune-zero} and \textit{prune-rewind} using three face attribute detectors (Figure \ref{fig:prune}). This parameter influences the degree of generalizability of the synthetic target representations. For \textit{prune-zero}, higher pruning ratios (severe style attenuation), \textit{i.e.}, $p$ between $80-90$, are found to significantly enhance performance when compared to lower ones. In the case of \textit{prune-rewind}, on the other hand, $p$ regulates the amount of source mix-up with the target domain. In this scenario, we see that a smaller $p$ performs better, and we recommend to set $p$ between $5-20$.


\textbf{\textbf{The choice of synthetic data size $T$.}}  We study the influence of the number of augmentations $T$ by varying it between $100-5000$ and studying the performance of \textit{prune-zero} and \textit{prune-rewind} on three attributes, as illustrated in Figure \ref{fig:varyT}. While \textit{prune-zero} performs consistently for different values of $T$, it only makes limited gains on average as the number of samples increases. On the contrary, we see a significant boost in performance in \textit{prune-rewind} in some of attributes. We remark that \textit{prune-rewind} is a sensitive technique due to the mix-up with the source domain; increasing the number of the synthetic augmentations (along with low $p$) stabilizes the performance and, in a few cases, even matches the performance of \textit{prune-zero}. Finally, we note that the performance variation across the independent trials is around $<0.5\%$, thus indicating that the performance is consistent and not sensitive to the sampling process.

\textbf{\textbf{Toolbox augmentations can further bolster \alg.}} 
In this study, we investigated the benefits of using sophisticated toolbox augmentations such as Augmix for \alg~as well as for the source only baseline. From Figure~\ref{fig:sista_augmix}, we observe a consistent boost in performance for all the three variants of \alg with average improvements of $6\%$, $4.2\%$ and almost $13.3\%$ respectively. These results highlight the effective complementary nature of \alg~to toolbox augmentations. Furthermore, it is worth noting that applying Augmix to the source-only methods does not lead to the same level of improvements. This observation is consistent with the findings from~\cite{cattan}, which noted that toolbox augmentations alone are insufficient to enhance adaptation performance under real-world distribution shifts.

\section{Conclusion}
In this paper, we explored the use of generative augmentations for test-time adaptation, when only a single-shot target is available. Through a combination of StyleGAN fine-tuning and novel sampling strategies, we were able to curate synthetic target datasets that effectively reflect the characteristics of any target domain.
We showed that the proposed approach is effective in multi-class classification using both class-conditioned as well as multiple class-specific GANs.
Our future work includes theoretically understanding the behavior of different pruning techniques and extending our approach beyond classifier adaptation.

\section*{Acknowledgements}
This work was performed under the auspices of the U.S. Department of Energy by the Lawrence Livermore National Laboratory under Contract No. DE-AC52-07NA27344. Supported by the LDRD Program under project 21-ERD-012. LLNL-CONF-844756.

\bibliography{main}

\begin{thebibliography}{46}
\providecommand{\natexlab}[1]{#1}
\providecommand{\url}[1]{\texttt{#1}}
\expandafter\ifx\csname urlstyle\endcsname\relax
  \providecommand{\doi}[1]{doi: #1}\else
  \providecommand{\doi}{doi: \begingroup \urlstyle{rm}\Url}\fi

\bibitem[Choi et~al.(2020)Choi, Uh, Yoo, and Ha]{choi2020starganv2}
Choi, Y., Uh, Y., Yoo, J., and Ha, J.-W.
\newblock Stargan v2: Diverse image synthesis for multiple domains.
\newblock In \emph{Proceedings of the IEEE Conference on Computer Vision and
  Pattern Recognition}, 2020.

\bibitem[Chong \& Forsyth(2021)Chong and Forsyth]{chong2021jojogan}
Chong, M.~J. and Forsyth, D.
\newblock Jojogan: One shot face stylization.
\newblock \emph{arXiv preprint arXiv:2112.11641}, 2021.

\bibitem[Cubuk et~al.(2019)Cubuk, Zoph, Mané, Vasudevan, and Le]{autoaugment1}
Cubuk, E.~D., Zoph, B., Mané, D., Vasudevan, V., and Le, Q.~V.
\newblock Autoaugment: Learning augmentation strategies from data.
\newblock In \emph{2019 IEEE/CVF Conference on Computer Vision and Pattern
  Recognition (CVPR)}, pp.\  113--123, 2019.
\newblock \doi{10.1109/CVPR.2019.00020}.

\bibitem[DeVries \& Taylor(2017)DeVries and Taylor]{cutout1}
DeVries, T. and Taylor, G.~W.
\newblock Improved regularization of convolutional neural networks with cutout.
\newblock \emph{arXiv preprint arXiv:1708.04552}, 2017.

\bibitem[Gokhale et~al.(2023)Gokhale, Anirudh, Thiagarajan, Kailkhura, Baral,
  and Yang]{gokhale2023improving}
Gokhale, T., Anirudh, R., Thiagarajan, J.~J., Kailkhura, B., Baral, C., and
  Yang, Y.
\newblock Improving diversity with adversarially learned transformations for
  domain generalization.
\newblock In \emph{Proceedings of the IEEE/CVF Winter Conference on
  Applications of Computer Vision}, pp.\  434--443, 2023.

\bibitem[He et~al.(2016)He, Zhang, Ren, and Sun]{he2016deep}
He, K., Zhang, X., Ren, S., and Sun, J.
\newblock Deep residual learning for image recognition.
\newblock In \emph{Proceedings of the IEEE conference on computer vision and
  pattern recognition}, pp.\  770--778, 2016.

\bibitem[Hendrycks \& Dietterich(2019)Hendrycks and
  Dietterich]{hendrycks2019robustness}
Hendrycks, D. and Dietterich, T.
\newblock Benchmarking neural network robustness to common corruptions and
  perturbations.
\newblock \emph{Proceedings of the International Conference on Learning
  Representations}, 2019.

\bibitem[Hendrycks et~al.(2020)Hendrycks, Mu, Cubuk, Zoph, Gilmer, and
  Lakshminarayanan]{hendrycks2019augmix}
Hendrycks, D., Mu, N., Cubuk, E.~D., Zoph, B., Gilmer, J., and
  Lakshminarayanan, B.
\newblock Augmix: {A} simple data processing method to improve robustness and
  uncertainty.
\newblock In \emph{8th International Conference on Learning Representations,
  {ICLR} 2020, Addis Ababa, Ethiopia, April 26-30, 2020}. OpenReview.net, 2020.
\newblock URL \url{https://openreview.net/forum?id=S1gmrxHFvB}.

\bibitem[Hendrycks et~al.(2021)]{hendrycks2021many}
Hendrycks, D. et~al.
\newblock The many faces of robustness: A critical analysis of
  out-of-distribution generalization.
\newblock In \emph{Proceedings of the IEEE/CVF International Conference on
  Computer Vision}, pp.\  8340--8349, 2021.

\bibitem[Hoffman et~al.(2018)]{hoffman2018cycada}
Hoffman, J. et~al.
\newblock Cycada: Cycle-consistent adversarial domain adaptation.
\newblock In \emph{International conference on machine learning}, pp.\
  1989--1998. Pmlr, 2018.

\bibitem[Huang et~al.(2021)Huang, Guan, Xiao, and Lu]{huang2021model}
Huang, J., Guan, D., Xiao, A., and Lu, S.
\newblock Model adaptation: Historical contrastive learning for unsupervised
  domain adaptation without source data.
\newblock \emph{Advances in Neural Information Processing Systems},
  34:\penalty0 3635--3649, 2021.

\bibitem[Ishii \& Sugiyama(2021)Ishii and Sugiyama]{ishii2021source}
Ishii, M. and Sugiyama, M.
\newblock Source-free domain adaptation via distributional alignment by
  matching batch normalization statistics.
\newblock \emph{arXiv preprint arXiv:2101.10842}, 2021.

\bibitem[Kafri et~al.(2021)Kafri, Patashnik, Alaluf, and
  Cohen-Or]{kafri2021stylefusion}
Kafri, O., Patashnik, O., Alaluf, Y., and Cohen-Or, D.
\newblock Stylefusion: A generative model for disentangling spatial segments.
\newblock \emph{arXiv preprint arXiv:2107.07437}, 2021.

\bibitem[Karras et~al.(2017)Karras, Aila, Laine, and
  Lehtinen]{DBLP:journals/corr/abs-1710-10196}
Karras, T., Aila, T., Laine, S., and Lehtinen, J.
\newblock Progressive growing of gans for improved quality, stability, and
  variation.
\newblock \emph{CoRR}, abs/1710.10196, 2017.
\newblock URL \url{http://arxiv.org/abs/1710.10196}.

\bibitem[Karras et~al.(2019)Karras, Laine, and Aila]{karras2019style}
Karras, T., Laine, S., and Aila, T.
\newblock A style-based generator architecture for generative adversarial
  networks.
\newblock In \emph{Proceedings of the IEEE/CVF conference on computer vision
  and pattern recognition}, pp.\  4401--4410, 2019.

\bibitem[Karras et~al.(2020)Karras, Laine, Aittala, Hellsten, Lehtinen, and
  Aila]{karras2020analyzing}
Karras, T., Laine, S., Aittala, M., Hellsten, J., Lehtinen, J., and Aila, T.
\newblock Analyzing and improving the image quality of stylegan.
\newblock In \emph{Proceedings of the IEEE/CVF conference on computer vision
  and pattern recognition}, pp.\  8110--8119, 2020.

\bibitem[Krizhevsky et~al.(2009)Krizhevsky, Hinton,
  et~al.]{krizhevsky2009learning}
Krizhevsky, A., Hinton, G., et~al.
\newblock Learning multiple layers of features from tiny images.
\newblock 2009.

\bibitem[Kundu et~al.(2020)Kundu, Venkat, Revanur, Babu,
  et~al.]{kundu2020towards}
Kundu, J.~N., Venkat, N., Revanur, A., Babu, R.~V., et~al.
\newblock Towards inheritable models for open-set domain adaptation.
\newblock In \emph{Proceedings of the IEEE/CVF Conference on Computer Vision
  and Pattern Recognition}, pp.\  12376--12385, 2020.

\bibitem[Liang et~al.()Liang, Hu, and Feng]{SHOT}
Liang, J., Hu, D., and Feng, J.
\newblock Do we really need to access the source data? {S}ource hypothesis
  transfer for unsupervised domain adaptation.
\newblock In \emph{Proceedings of the 37th International Conference on Machine
  Learning}.

\bibitem[Liu \& Yuan(2022)Liu and Yuan]{liu2022source}
Liu, X. and Yuan, Y.
\newblock A source-free domain adaptive polyp detection framework with style
  diversification flow.
\newblock \emph{IEEE Transactions on Medical Imaging}, 41\penalty0
  (7):\penalty0 1897--1908, 2022.

\bibitem[Patashnik et~al.(2021)Patashnik, Wu, Shechtman, Cohen-Or, and
  Lischinski]{patashnik2021styleclip}
Patashnik, O., Wu, Z., Shechtman, E., Cohen-Or, D., and Lischinski, D.
\newblock Styleclip: Text-driven manipulation of stylegan imagery.
\newblock In \emph{Proceedings of the IEEE/CVF International Conference on
  Computer Vision}, pp.\  2085--2094, 2021.

\bibitem[Peng et~al.(2019)Peng, Bai, Xia, Huang, Saenko, and Wang]{domainnet}
Peng, X., Bai, Q., Xia, X., Huang, Z., Saenko, K., and Wang, B.
\newblock Moment matching for multi-source domain adaptation.
\newblock In \emph{Proceedings of the IEEE/CVF international conference on
  computer vision}, pp.\  1406--1415, 2019.

\bibitem[Rahman et~al.(2021)Rahman, Rahman, and Mahdy]{Rahman20213CGANCC}
Rahman, A., Rahman, M.~S., and Mahdy, M. R.~C.
\newblock 3c-gan: class-consistent cyclegan for malaria domain adaptation
  model.
\newblock \emph{Biomedical Physics \& Engineering Express}, 7, 2021.

\bibitem[Richardson et~al.(2021)]{richardson2021encoding}
Richardson, E. et~al.
\newblock Encoding in style: a stylegan encoder for image-to-image translation.
\newblock In \emph{Proceedings of the IEEE/CVF conference on computer vision
  and pattern recognition}, pp.\  2287--2296, 2021.

\bibitem[Robey et~al.(2021)Robey, Pappas, and Hassani]{robey2021model}
Robey, A., Pappas, G.~J., and Hassani, H.
\newblock Model-based domain generalization.
\newblock \emph{Advances in Neural Information Processing Systems},
  34:\penalty0 20210--20229, 2021.

\bibitem[Russakovsky et~al.(2015)Russakovsky, Deng, Su, Krause, Satheesh, Ma,
  Huang, Karpathy, Khosla, Bernstein, Berg, and Fei-Fei]{ILSVRC15}
Russakovsky, O., Deng, J., Su, H., Krause, J., Satheesh, S., Ma, S., Huang, Z.,
  Karpathy, A., Khosla, A., Bernstein, M., Berg, A.~C., and Fei-Fei, L.
\newblock {ImageNet Large Scale Visual Recognition Challenge}.
\newblock \emph{International Journal of Computer Vision (IJCV)}, 115\penalty0
  (3):\penalty0 211--252, 2015.
\newblock \doi{10.1007/s11263-015-0816-y}.

\bibitem[Saharia et~al.(2022)Saharia, Chan, Saxena, Li, Whang, Denton,
  Ghasemipour, Ayan, Mahdavi, Lopes, et~al.]{saharia2022photorealistic}
Saharia, C., Chan, W., Saxena, S., Li, L., Whang, J., Denton, E., Ghasemipour,
  S. K.~S., Ayan, B.~K., Mahdavi, S.~S., Lopes, R.~G., et~al.
\newblock Photorealistic text-to-image diffusion models with deep language
  understanding.
\newblock \emph{arXiv preprint arXiv:2205.11487}, 2022.

\bibitem[Sankaranarayanan et~al.(2018)Sankaranarayanan, Balaji, Castillo, and
  Chellappa]{sankaranarayanan2018generate}
Sankaranarayanan, S., Balaji, Y., Castillo, C.~D., and Chellappa, R.
\newblock Generate to adapt: Aligning domains using generative adversarial
  networks.
\newblock In \emph{Proceedings of the IEEE conference on computer vision and
  pattern recognition}, pp.\  8503--8512, 2018.

\bibitem[Sauer et~al.(2022)Sauer, Schwarz, and Geiger]{sauer2022stylegan}
Sauer, A., Schwarz, K., and Geiger, A.
\newblock Stylegan-xl: Scaling stylegan to large diverse datasets.
\newblock In \emph{ACM SIGGRAPH 2022 conference proceedings}, pp.\  1--10,
  2022.

\bibitem[Steiner et~al.(2021)Steiner, Kolesnikov, Zhai, Wightman, Uszkoreit,
  and Beyer]{steiner2021train}
Steiner, A., Kolesnikov, A., Zhai, X., Wightman, R., Uszkoreit, J., and Beyer,
  L.
\newblock How to train your vit? data, augmentation, and regularization in
  vision transformers.
\newblock \emph{arXiv preprint arXiv:2106.10270}, 2021.

\bibitem[Subramanyam et~al.(2022)Subramanyam, Narayanaswamy, Naufel, Spanias,
  and Thiagarajan]{subramanyam2022improved}
Subramanyam, R., Narayanaswamy, V., Naufel, M., Spanias, A., and Thiagarajan,
  J.~J.
\newblock Improved stylegan-v2 based inversion for out-of-distribution images.
\newblock In \emph{International Conference on Machine Learning}, pp.\
  20625--20639. PMLR, 2022.

\bibitem[Sun et~al.(2020)Sun, Wang, Liu, Miller, Efros, and Hardt]{sun2020test}
Sun, Y., Wang, X., Liu, Z., Miller, J., Efros, A., and Hardt, M.
\newblock Test-time training with self-supervision for generalization under
  distribution shifts.
\newblock In \emph{International conference on machine learning}, pp.\
  9229--9248. PMLR, 2020.

\bibitem[Tang et~al.(2021)Tang, Yang, Ma, Hendrich, Zeng, Ge, Zhang, and
  Zhang]{tang2021nearest}
Tang, S., Yang, Y., Ma, Z., Hendrich, N., Zeng, F., Ge, S.~S., Zhang, C., and
  Zhang, J.
\newblock Nearest neighborhood-based deep clustering for source data-absent
  unsupervised domain adaptation.
\newblock \emph{arXiv preprint arXiv:2107.12585}, 2021.

\bibitem[Thopalli et~al.(2022)Thopalli, Turaga, and Thiagarajan]{cattan}
Thopalli, K., Turaga, P., and Thiagarajan, J.~J.
\newblock Domain alignment meets fully test-time adaptation.
\newblock In \emph{Asian Conference on Machine Learning, 2022.}, 2022.

\bibitem[Torralba \& Efros(2011)Torralba and Efros]{torralba2011unbiased}
Torralba, A. and Efros, A.~A.
\newblock Unbiased look at dataset bias.
\newblock In \emph{CVPR 2011}, pp.\  1521--1528. IEEE, 2011.

\bibitem[Tov et~al.(2021)Tov, Alaluf, Nitzan, Patashnik, and
  Cohen-Or]{tov2021designing}
Tov, O., Alaluf, Y., Nitzan, Y., Patashnik, O., and Cohen-Or, D.
\newblock Designing an encoder for stylegan image manipulation.
\newblock \emph{ACM Transactions on Graphics (TOG)}, 40\penalty0 (4):\penalty0
  1--14, 2021.

\bibitem[Wang et~al.(2021)Wang, Shelhamer, Liu, Olshausen, and Darrell]{Tent}
Wang, D., Shelhamer, E., Liu, S., Olshausen, B., and Darrell, T.
\newblock Tent: Fully test-time adaptation by entropy minimization.
\newblock In \emph{International Conference on Learning Representations}, 2021.
\newblock URL \url{https://openreview.net/forum?id=uXl3bZLkr3c}.

\bibitem[Wu et~al.(2021{\natexlab{a}})Wu, Lischinski, and
  Shechtman]{wu2021stylespace}
Wu, Z., Lischinski, D., and Shechtman, E.
\newblock Stylespace analysis: Disentangled controls for stylegan image
  generation.
\newblock In \emph{Proceedings of the IEEE/CVF Conference on Computer Vision
  and Pattern Recognition}, pp.\  12863--12872, 2021{\natexlab{a}}.

\bibitem[Wu et~al.(2021{\natexlab{b}})Wu, Nitzan, Shechtman, and
  Lischinski]{wu2021stylealign}
Wu, Z., Nitzan, Y., Shechtman, E., and Lischinski, D.
\newblock Stylealign: Analysis and applications of aligned stylegan models.
\newblock \emph{arXiv preprint arXiv:2110.11323}, 2021{\natexlab{b}}.

\bibitem[Xia et~al.(2022)Xia, Zhang, Yang, Xue, Zhou, and Yang]{xia2022gan}
Xia, W., Zhang, Y., Yang, Y., Xue, J.-H., Zhou, B., and Yang, M.-H.
\newblock Gan inversion: A survey.
\newblock \emph{IEEE Transactions on Pattern Analysis and Machine
  Intelligence}, 2022.

\bibitem[Xu et~al.(2021)Xu, Liu, Yang, Raffel, and Niethammer]{randconv}
Xu, Z., Liu, D., Yang, J., Raffel, C., and Niethammer, M.
\newblock Robust and generalizable visual representation learning via random
  convolutions.
\newblock In \emph{International Conference on Learning Representations}, 2021.
\newblock URL \url{https://openreview.net/forum?id=BVSM0x3EDK6}.

\bibitem[Yang et~al.(2021)Yang, van~de Weijer, Herranz, Jui,
  et~al.]{yang2021exploiting}
Yang, S., van~de Weijer, J., Herranz, L., Jui, S., et~al.
\newblock Exploiting the intrinsic neighborhood structure for source-free
  domain adaptation.
\newblock \emph{Advances in Neural Information Processing Systems},
  34:\penalty0 29393--29405, 2021.

\bibitem[Yue et~al.(2022)Yue, Zhang, Yuan, Xu, and Song]{SurveyGanAug}
Yue, F., Zhang, C., Yuan, M., Xu, C., and Song, Y.
\newblock Survey of image augmentation based on generative adversarial network.
\newblock \emph{Journal of Physics: Conference Series}, 2203\penalty0
  (1):\penalty0 012052, feb 2022.
\newblock \doi{10.1088/1742-6596/2203/1/012052}.
\newblock URL \url{https://dx.doi.org/10.1088/1742-6596/2203/1/012052}.

\bibitem[Yun et~al.(2019)Yun, Han, Oh, Chun, Choe, and Yoo]{cutmix2}
Yun, S., Han, D., Oh, S.~J., Chun, S., Choe, J., and Yoo, Y.
\newblock Cutmix: Regularization strategy to train strong classifiers with
  localizable features.
\newblock In \emph{Proceedings of the IEEE/CVF international conference on
  computer vision}, pp.\  6023--6032, 2019.

\bibitem[Zhang et~al.(2018)Zhang, Cisse, Dauphin, and Lopez-Paz]{mixup}
Zhang, H., Cisse, M., Dauphin, Y.~N., and Lopez-Paz, D.
\newblock mixup: Beyond empirical risk minimization.
\newblock In \emph{International Conference on Learning Representations}, 2018.

\bibitem[Zhang et~al.(2021)Zhang, Levine, and Finn]{zhang2021memo}
Zhang, M., Levine, S., and Finn, C.
\newblock Memo: Test time robustness via adaptation and augmentation.
\newblock \emph{arXiv preprint arXiv:2110.09506}, 2021.

\end{thebibliography}
\bibliographystyle{icml2023}

\newpage
\appendix
    
\section{Examples of augmentations from \alg}
\begin{figure}[h]
    \centering
    \includegraphics[width=0.95\columnwidth]{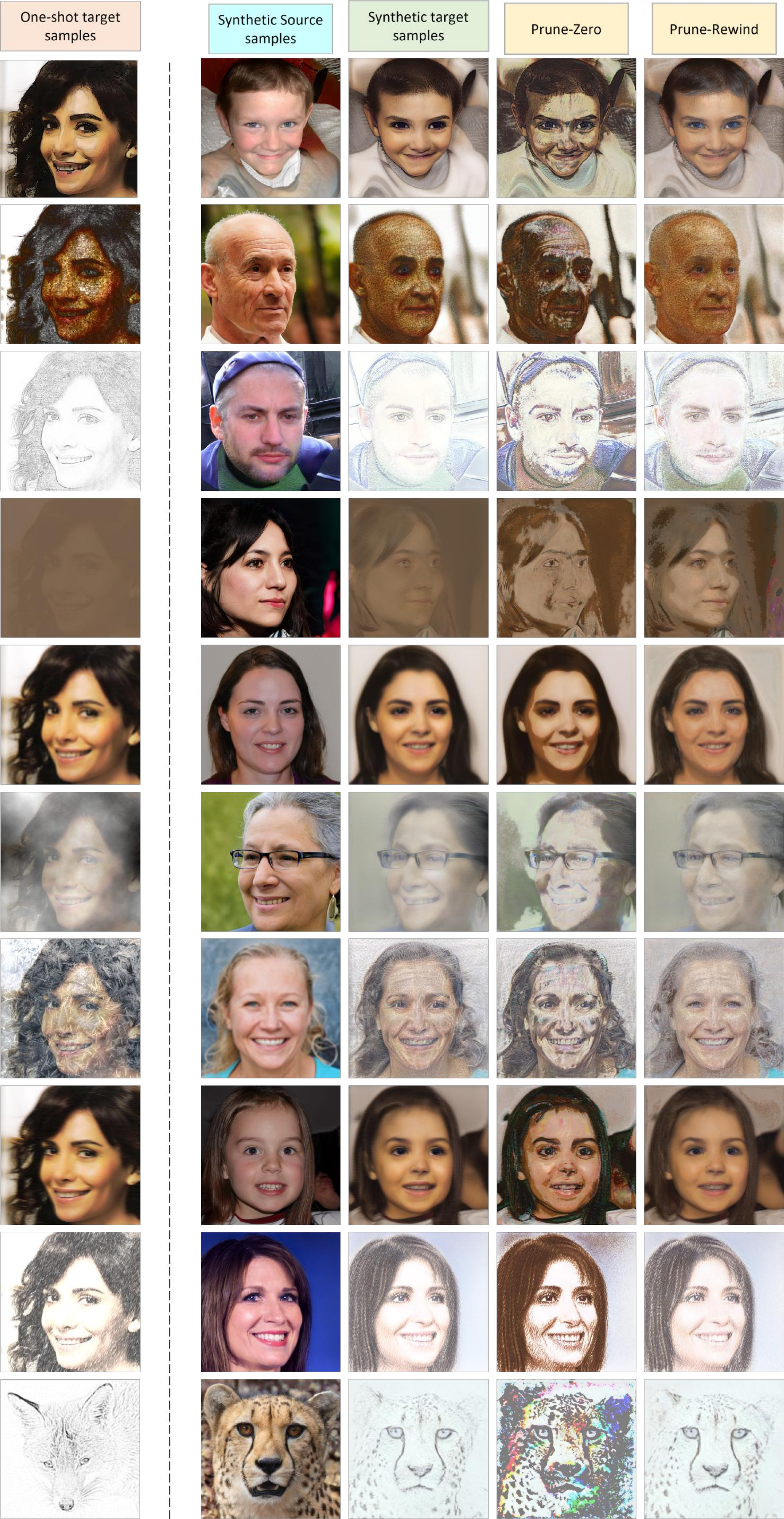}
    \caption{\textbf{\alg~generated augmentations} on random samples drawn from the style space of StyleGAN; The rows 1 to 9 correspond to different domain shifts in CelebA-HQ and row 10 corresponds to AFHQ.}
    \label{fig:domain_augs}
\end{figure}

In Figure \ref{fig:domain_augs}, we show the augmentations synthesized by \alg~for different domain shifts and StyleGAN models.

\newpage

\newpage

\section{Detailed results for our CelebA experiments}
We provide comprehensive tables for the results discussed in Section \ref{sec:results}. Tables \ref{tab:DomainA}-\ref{tab:contrast} illustrate the performance of source-only, MEMO, and all the three variants of \alg~along with Full target performance.

\begin{table*}[h]
\renewcommand{\arraystretch}{1.5}
\resizebox{0.99\linewidth}{!}{
\begin{tabular}{|l|c|c|c|c|c|c|c|c|c|c|c|c|c|c|c|c|c|c|c|}
\cline{2-20}
\multicolumn{1}{c|}{}                                 & {\rotatebox[origin=c]{90}{5'o clock shadow}} & \rotatebox[origin=c]{90}{Arched eyebrows} & \rotatebox[origin=c]{90}{Bald} & \rotatebox[origin=c]{90}{Bangs} & \rotatebox[origin=c]{90}{Blond hair} & \rotatebox[origin=c]{90}{Eyeglasses} & \rotatebox[origin=c]{90}{Makeup} & \rotatebox[origin=c]{90}{Cheekbones} & \rotatebox[origin=c]{90}{Gender} & \rotatebox[origin=c]{90}{Mouth open} & \rotatebox[origin=c]{90}{Eyes closed} & \rotatebox[origin=c]{90}{Beard} & \rotatebox[origin=c]{90}{Sideburns} & \rotatebox[origin=c]{90}{Smiling} & \rotatebox[origin=c]{90}{Straight hair} & \rotatebox[origin=c]{90}{Wavy hair} & \rotatebox[origin=c]{90}{Earrings} & \rotatebox[origin=c]{90}{Lipstick} & \rotatebox[origin=c]{90}{Young} \\
\cline{2-20} \hline
\cellcolor[HTML]{FFE6E6} Source only &53.7&69.9&63.4&83.2&55.3&\textbf{89.5}&80.7&80.4&93.2&88.2&58.1&\textbf{82}&60.2&\textbf{89.5}&53.4&68.3&70.9&88.5&64.8 \\ \hline \hline
MEMO (Augmix) &53.6&69.9&64.5&81.1&53.8&89.1&79.7&78.6&93.8&87.6&57.9&80.8&59.6&89.4&52.5&70.5&68.6&88.1&65.1 \\ \hline
MEMO (Randconv) &53.7&69.6&64.5&81&53.7&89.1&79.5&78.4&93.9&87.6&57.9&80.8&59.5&89.3&52.5&70.2&68.6&88&65 \\ \hline \hline
SiSTA (base) &52.8&74.6&\textbf{77}&80&85.2&69.8&87.2&72.8&95.1&91.2&55.2&69.8&58.3&84.4&57&79.1&71.3&\textbf{90.1}&\textbf{69.1} \\ \hline
SiSTA (prune-zero)&\textbf{55.2}&\textbf{78.2}&76.3&\textbf{87.1}&\textbf{87.6}&81.5&\textbf{88.1}&\textbf{81.2}&\textbf{95.5}&\textbf{91.7}&\textbf{60.4}&70.8&\textbf{61.1}&89.2&\textbf{59.3}&\textbf{79.5}&\textbf{76.2}&89.6&68.6 \\ \hline
SiSTA (prune-rewind)&53.1&76.6&70.1&85.6&83&78.2&87.1&76&95.2&91.6&57.8&67.5&58.5&87.3&59.2&78.6&74.2&89.3&60.6 \\ \hline \hline
\cellcolor[HTML]{E6FFF2} Full target DA &87&81.9&92.3&93.5&90.1&97.3&89.3&87.1&97.4&92.7&72.5&91.5&93&92.6&74.5&80.6&82.5&92.3&75.2 \\ \hline
\end{tabular}
}
\caption{Performance of \alg~on Domain A of the CelebA dataset.}
\label{tab:DomainA}
\end{table*}

\begin{table*}[th]
\renewcommand{\arraystretch}{1.5}
\resizebox{0.99\linewidth}{!}{
\begin{tabular}{|l|c|c|c|c|c|c|c|c|c|c|c|c|c|c|c|c|c|c|c|}
\cline{2-20}
\multicolumn{1}{c|}{}                                 & {\rotatebox[origin=c]{90}{\textcolor{red}{5'o clock shadow}}} & \rotatebox[origin=c]{90}{Arched eyebrows} & \rotatebox[origin=c]{90}{\textcolor{red}{Bald}} & \rotatebox[origin=c]{90}{Bangs} & \rotatebox[origin=c]{90}{\textcolor{red}{Blond hair}} & \rotatebox[origin=c]{90}{Eyeglasses} & \rotatebox[origin=c]{90}{Makeup} & \rotatebox[origin=c]{90}{Cheekbones} & \rotatebox[origin=c]{90}{Gender} & \rotatebox[origin=c]{90}{Mouth open} & \rotatebox[origin=c]{90}{Eyes closed} & \rotatebox[origin=c]{90}{Beard} & \rotatebox[origin=c]{90}{\textcolor{red}{Sideburns}} & \rotatebox[origin=c]{90}{Smiling} & \rotatebox[origin=c]{90}{Straight hair} & \rotatebox[origin=c]{90}{Wavy hair} & \rotatebox[origin=c]{90}{Earrings} & \rotatebox[origin=c]{90}{Lipstick} & \rotatebox[origin=c]{90}{Young} \\
\cline{2-20} \hline
\cellcolor[HTML]{FFE6E6} Source only &50&51&50.5&67.2&50&74.2&54.2&54.6&80.2&78.6&52.1&\textbf{63.9}&\textbf{54}&76.9&50.1&65&50.4&63.3&55.5 \\ \hline \hline
MEMO (Augmix) &50&51.2&50.5&64.5&50&74.1&52.1&52.4&81.1&79&51.2&63&50.8&73.2&50&65.5&50.2&58.6&55.6 \\ \hline
MEMO (Randconv) &50&51.2&50.5&64.5&50&73.9&52.1&52.3&81.2&79&51.2&62.9&50.8&73.1&50&65.6&50.2&58.5&55.7 \\ \hline \hline
SiSTA (base) &50&73&50.2&83.3&50.5&67.8&77.6&56.3&86.5&82.5&56.7&56.1&50.1&77&51.7&72.6&\textbf{56.3}&\textbf{80}&58.1 \\ \hline
SiSTA (prune-zero)&50.1&\textbf{73.9}&51.1&\textbf{86.7}&51.4&\textbf{75.8}&\textbf{79.9}&\textbf{67.2}&\textbf{88.7}&\textbf{84.4}&\textbf{58.3}&58.1&50.2&\textbf{85.4}&\textbf{53.8}&\textbf{74}&54.8&79.8&\textbf{60.5} \\ \hline
SiSTA (prune-rewind)&50&73.4&50&84.7&50.2&75.2&75.5&57.1&85.9&82.9&54&54.5&50.1&78&52.7&72.8&\textbf{56.3}&73&56.3 \\ \hline \hline
\cellcolor[HTML]{E6FFF2}Full target DA &71.6&71.7&72.6&89.9&58.4&94.2&81.9&78.5&92.2&88&63.9&84.3&83&88.4&68.6&71&68.6&86.7&71.2 \\ \hline
\end{tabular}
}
\caption{Performance of \alg~on Domain B of the CelebA dataset.}
\label{tab:DomainB}
\end{table*}

\begin{table*}[]
\renewcommand{\arraystretch}{1.5}
\resizebox{0.99\linewidth}{!}{
\begin{tabular}{|l|c|c|c|c|c|c|c|c|c|c|c|c|c|c|c|c|c|c|c|}
\cline{2-20}
\multicolumn{1}{c|}{}                                 & {\rotatebox[origin=c]{90}{5'o clock shadow}} & \rotatebox[origin=c]{90}{Arched eyebrows} & \rotatebox[origin=c]{90}{Bald} & \rotatebox[origin=c]{90}{Bangs} & \rotatebox[origin=c]{90}{Blond hair} & \rotatebox[origin=c]{90}{Eyeglasses} & \rotatebox[origin=c]{90}{Makeup} & \rotatebox[origin=c]{90}{Cheekbones} & \rotatebox[origin=c]{90}{Gender} & \rotatebox[origin=c]{90}{Mouth open} & \rotatebox[origin=c]{90}{Eyes closed} & \rotatebox[origin=c]{90}{Beard} & \rotatebox[origin=c]{90}{Sideburns} & \rotatebox[origin=c]{90}{Smiling} & \rotatebox[origin=c]{90}{Straight hair} & \rotatebox[origin=c]{90}{Wavy hair} & \rotatebox[origin=c]{90}{Earrings} & \rotatebox[origin=c]{90}{Lipstick} & \rotatebox[origin=c]{90}{Young} \\

\cline{2-20} \hline
\cellcolor[HTML]{FFE6E6} Source only &50&52.8&50.1&58.2&50.5&63.8&56.5&50.2&58.3&58.9&50&51.3&50.5&64&52&59.9&51.8&71.6&52.7 \\ \hline \hline
MEMO (Augmix) &50&53.6&50.2&61.6&50.5&66.6&55.5&50.1&56.1&60.4&50&50.8&50.4&65.8&52&59.2&51.7&72.3&52 \\ \hline
MEMO (Randconv) &50&53.6&50.2&61.6&50.5&66.6&55.4&50.1&56&60.4&50&50.7&50.4&65.5&52&59.1&51.7&72.4&52 \\ \hline \hline
SiSTA (base) &53.2&65.3&\textbf{64.7}&80&77.9&69.4&54.5&71.2&91.8&71.4&\textbf{59.1}&66.6&53.2&79.2&54.7&77.3&57.8&78.8&63.7 \\ \hline
SiSTA (prune-zero)&\textbf{58}&\textbf{74.7}&64.1&82.6&77.1&\textbf{82.7}&\textbf{80.7}&\textbf{77.2}&88.3&\textbf{78.2}&56.3&\textbf{68.2}&\textbf{55.3}&\textbf{86.7}&\textbf{68.5}&74.3&\textbf{62.8}&\textbf{86.5}&67.6 \\ \hline
SiSTA (prune-rewind)&53.1&69.7&63.5&\textbf{84.3}&\textbf{80.1}&79.9&62.1&69.7&\textbf{92.2}&\textbf{78.2}&54.4&65&53.7&84.4&57.3&\textbf{78.5}&58.2&\textbf{86.5}&\textbf{74.5} \\ \hline \hline
\cellcolor[HTML]{E6FFF2}  Full target DA &83.1&80.5&92&93&84.2&96.7&83.8&80.8&95.7&87.6&66.9&90&93.2&89.2&69.9&77.5&76.6&89.5&77.5 \\ \hline

\end{tabular}
}
\caption{Performance of \alg~on Domain C of the CelebA dataset.}
\label{tab:DomainC}
\end{table*}

\begin{table*}[]
\renewcommand{\arraystretch}{1.5}
\resizebox{0.99\linewidth}{!}{
\begin{tabular}{|l|c|c|c|c|c|c|c|c|c|c|c|c|c|c|c|c|c|c|c|}
\cline{2-20}
\multicolumn{1}{c|}{}                                 & {\rotatebox[origin=c]{90}{5'o clock shadow}} & \rotatebox[origin=c]{90}{Arched eyebrows} & \rotatebox[origin=c]{90}{Bald} & \rotatebox[origin=c]{90}{Bangs} & \rotatebox[origin=c]{90}{Blond hair} & \rotatebox[origin=c]{90}{Eyeglasses} & \rotatebox[origin=c]{90}{Makeup} & \rotatebox[origin=c]{90}{Cheekbones} & \rotatebox[origin=c]{90}{Gender} & \rotatebox[origin=c]{90}{Mouth open} & \rotatebox[origin=c]{90}{Eyes closed} & \rotatebox[origin=c]{90}{Beard} & \rotatebox[origin=c]{90}{Sideburns} & \rotatebox[origin=c]{90}{Smiling} & \rotatebox[origin=c]{90}{Straight hair} & \rotatebox[origin=c]{90}{Wavy hair} & \rotatebox[origin=c]{90}{Earrings} & \rotatebox[origin=c]{90}{Lipstick} & \rotatebox[origin=c]{90}{Young} \\
\cline{2-20} \hline
\cellcolor[HTML]{FFE6E6} Source only &64.5&79&82.5&90&87.4&91&90.4&87.8&97.2&92&64.5&79.7&63.4&93&68.8&79.9&65.7&92.6&74.9 \\ \hline\hline
MEMO (Augmix) &63.2&78.1&87.5&88&87.1&\textbf{91.3}&90.6&\textbf{89.8}&\textbf{97.8}&90.8&65.3&77.4&62&\textbf{92.9}&70.6&80.8&63.9&91&75.5 \\ \hline
MEMO (Randconv) &63.2&78.1&87.5&87.5&87.1&\textbf{91.3}&90.6&\textbf{89.8}&\textbf{97.8}&90.8&65.3&77.4&62&\textbf{92.9}&70.6&81&63.7&91&75.3 \\ \hline\hline
SiSTA (base) &\textbf{85.6}&80&\textbf{88.9}&88.9&91.2&76.9&89.8&79&95.3&91.5&\textbf{65.6}&\textbf{91.4}&\textbf{89.3}&87.5&65.2&\textbf{82.4}&68.2&91.9&\textbf{81.9}\\ \hline
SiSTA (prune-zero)&85.1&79.5&85.1&90.3&\textbf{92.8}&83.3&\textbf{90.7}&82.4&96.4&{90.7}&63.8&89.7&76.7&89.9&\textbf{73.7}&81.5&69.3&\textbf{92.2}&73.3\\ \hline
SiSTA (prune-rewind)&78.2&\textbf{81.5}&85.3&\textbf{92.3}&92.5&83.4&90.5&81.7&97.2&\textbf{92.7}&64.2&87.7&77.3&90.7&71.1&82.1&\textbf{71.1}&\textbf{92.2}&75.5\\ \hline\hline
 \cellcolor[HTML]{E6FFF2} Full target DA &89.4&83&96.1&94&92.9&97.1&90.7&88&97.8&93.7&74.4&93.3&94.1&93.3&76.9&82.4&84.5&92.6&83.1\\ \hline
\end{tabular}
}
\caption{Performance of \alg~on Domain D (Defocus blur) of the CelebA dataset.}
\label{tab:defocus}
\end{table*}

\begin{table*}[]
\renewcommand{\arraystretch}{1.5}
\resizebox{0.99\linewidth}{!}{
\begin{tabular}{|l|c|c|c|c|c|c|c|c|c|c|c|c|c|c|c|c|c|c|c|}
\cline{2-20}
\multicolumn{1}{c|}{}                                 & {\rotatebox[origin=c]{90}{5'o clock shadow}} & \rotatebox[origin=c]{90}{Arched eyebrows} & \rotatebox[origin=c]{90}{Bald} & \rotatebox[origin=c]{90}{Bangs} & \rotatebox[origin=c]{90}{Blond hair} & \rotatebox[origin=c]{90}{Eyeglasses} & \rotatebox[origin=c]{90}{Makeup} & \rotatebox[origin=c]{90}{Cheekbones} & \rotatebox[origin=c]{90}{Gender} & \rotatebox[origin=c]{90}{Mouth open} & \rotatebox[origin=c]{90}{Eyes closed} & \rotatebox[origin=c]{90}{Beard} & \rotatebox[origin=c]{90}{Sideburns} & \rotatebox[origin=c]{90}{Smiling} & \rotatebox[origin=c]{90}{Straight hair} & \rotatebox[origin=c]{90}{Wavy hair} & \rotatebox[origin=c]{90}{Earrings} & \rotatebox[origin=c]{90}{Lipstick} & \rotatebox[origin=c]{90}{Young} \\

\cline{2-20} \hline
 \cellcolor[HTML]{FFE6E6}Source only &71.4&\textbf{79.7}&79.5&88.3&88.9&91.5&89.7&87.1&97.6&91.6&69.6&80.5&65.7&92.9&72.5&73.9&62.2&92.2&74.8 \\ \hline \hline
MEMO (Augmix) &73&78.6&73.7&88.3&88.8&\textbf{91.8}&91.9&\textbf{88.5}&\textbf{97.5}&92&\textbf{70.7}&80.8&63&\textbf{93.1}&\textbf{73.5}&75&62.2&92.6&75.5 \\ \hline
MEMO (Randconv) &73&78.6&73.7&88.3&88.8&\textbf{91.8}&\textbf{92}&\textbf{88.5}&\textbf{97.5}&92.1&\textbf{70.7}&80.8&63&\textbf{93.1}&\textbf{73.5}&75&62.2&\textbf{92.7}&75.5 \\ \hline \hline
SiSTA (base) &\textbf{79.8}&74.7&\textbf{89.8}&89.3&\textbf{93.6}&78.2&89.6&79.5&94.4&92.2&67.4&\textbf{87.8}&\textbf{73.1}&87.9&69.7&\textbf{81.5}&\textbf{71}&92&\textbf{82}\\ \hline
SiSTA (prune-zero)&74&75.4&87.1&92.1&\textbf{93.6}&86.9&90.6&83.7&96.5&91.4&66.3&78.6&63&90.8&72.9&81.2&70.9&92.4&76.3 \\ \hline
SiSTA (prune-rewind)&70.7&76.1&85.9&\textbf{92.5}&\textbf{93.6}&85.5&90&81.2&96.2&\textbf{92.8}&65.9&79.7&64.9&89.9&72.5&80.4&68.9&92.2&73.9 \\ \hline \hline
 \cellcolor[HTML]{E6FFF2} Full target DA &90.1&82.8&96.7&93.8&93.2&98.1&90.8&88.2&97.9&93.7&72&94.9&94.6&93.2&75.7&82.6&85.4&92.9&84.2 \\ \hline
\end{tabular}
}
\caption{Performance of \alg~on Domain D (Motion blur) of the CelebA dataset.}
\label{tab:motion}
\end{table*}

\begin{table*}[]
\renewcommand{\arraystretch}{1.5}
\resizebox{0.99\linewidth}{!}{
\begin{tabular}{|l|c|c|c|c|c|c|c|c|c|c|c|c|c|c|c|c|c|c|c|}
\cline{2-20}
\multicolumn{1}{c|}{}                                 & {\rotatebox[origin=c]{90}{5'o clock shadow}} & \rotatebox[origin=c]{90}{Arched eyebrows} & \rotatebox[origin=c]{90}{Bald} & \rotatebox[origin=c]{90}{Bangs} & \rotatebox[origin=c]{90}{Blond hair} & \rotatebox[origin=c]{90}{Eyeglasses} & \rotatebox[origin=c]{90}{Makeup} & \rotatebox[origin=c]{90}{Cheekbones} & \rotatebox[origin=c]{90}{Gender} & \rotatebox[origin=c]{90}{Mouth open} & \rotatebox[origin=c]{90}{Eyes closed} & \rotatebox[origin=c]{90}{Beard} & \rotatebox[origin=c]{90}{Sideburns} & \rotatebox[origin=c]{90}{Smiling} & \rotatebox[origin=c]{90}{Straight hair} & \rotatebox[origin=c]{90}{Wavy hair} & \rotatebox[origin=c]{90}{Earrings} & \rotatebox[origin=c]{90}{Lipstick} & \rotatebox[origin=c]{90}{Young} \\

\cline{2-20} \hline
 \cellcolor[HTML]{FFE6E6}Source only &59.5&52.5&\textbf{71.9}&59.9&51.9&88.1&50&57.6&78.3&79.6&50.6&82&63&77.2&\textbf{53.8}&63.6&52.5&50.6&62.5\\ \hline\hline
MEMO (Augmix) &\textbf{62.6}&52.5&60.9&61.4&52.4&83&50&57.9&78.2&78.9&50.6&81.3&\textbf{64.1}&77.1&52&62.3&53.1&50.5&61.1\\ \hline
MEMO (Randconv) &\textbf{62.6}&52.4&60.9&61.1&52.4&83&50&57.9&78.3&78.9&50.6&81.3&63.5&77&51.6&62.3&53.1&50.5&61.1\\ \hline\hline
SiSTA (base) &57.3&56.3&\textbf{71.9}&77.9&57.4&80.6&60.7&68.2&75.2&84.2&\textbf{57.1}&\textbf{84.9}&63&76.8&51.8&74.8&62.3&73.5&69.6\\ \hline
SiSTA (prune-zero)&54.1&57.3&70.8&80.6&\textbf{58.9}&\textbf{89}&63.6&\textbf{77}&\textbf{81.1}&82.8&56.4&73.9&55.4&\textbf{85.3}&53.7&\textbf{76.2}&\textbf{63.8}&78.2&\textbf{71.7}\\ \hline
SiSTA (prune-rewind)&54.3&\textbf{58.5}&68.8&\textbf{84.3}&53.6&87.3&\textbf{69.4}&75.9&78.4&\textbf{85.8}&56.1&80.9&60&81.5&52.1&74.8&62.2&\textbf{80.8}&70.6\\ \hline\hline
 \cellcolor[HTML]{E6FFF2} Full target DA &86.9&78.8&80&90.6&90&97.8&85.9&82.9&94.6&92.9&72.6&92.6&92.5&89&70.6&78.3&84.2&89.6&76\\ \hline
\end{tabular}
}
\caption{Performance of \alg~on Domain D (Fog) of the CelebA dataset.}
\label{tab:fog}
\end{table*}

\begin{table*}[]
\renewcommand{\arraystretch}{1.5}
\resizebox{0.99\linewidth}{!}{
\begin{tabular}{|l|c|c|c|c|c|c|c|c|c|c|c|c|c|c|c|c|c|c|c|}
\cline{2-20}
\multicolumn{1}{c|}{}                                 & {\rotatebox[origin=c]{90}{\textcolor{red}{5'o clock shadow}}} & \rotatebox[origin=c]{90}{Arched eyebrows} & \rotatebox[origin=c]{90}{Bald} & \rotatebox[origin=c]{90}{Bangs} & \rotatebox[origin=c]{90}{Blond hair} & \rotatebox[origin=c]{90}{Eyeglasses} & \rotatebox[origin=c]{90}{Makeup} & \rotatebox[origin=c]{90}{Cheekbones} & \rotatebox[origin=c]{90}{Gender} & \rotatebox[origin=c]{90}{Mouth open} & \rotatebox[origin=c]{90}{Eyes closed} & \rotatebox[origin=c]{90}{Beard} & \rotatebox[origin=c]{90}{\textcolor{red}{Sideburns}} & \rotatebox[origin=c]{90}{Smiling} & \rotatebox[origin=c]{90}{Straight hair} & \rotatebox[origin=c]{90}{Wavy hair} & \rotatebox[origin=c]{90}{Earrings} & \rotatebox[origin=c]{90}{Lipstick} & \rotatebox[origin=c]{90}{Young} \\

\cline{2-20} \hline
 \cellcolor[HTML]{FFE6E6}Source only &51.1&51.5&54.6&55.8&51.3&70.5&50.1&53.5&75.5&72.9&50.8&68.6&56.3&66.7&50.2&64.6&51.7&51.5&54.7 \\ \hline\hline
MEMO (Augmix) &50.3&51.4&55.6&55.8&51.4&\textbf{72}&50.2&53.9&75.9&74.3&51.1&68.3&56.4&67.2&50&64.3&51.2&51.1&53.9\\ \hline
MEMO (Randconv) &50.3&51.4&55.6&55.8&51.4&71&50.2&53.9&75.9&74.4&51.1&68.4&56.4&67.2&50&64&51.2&51.1&54\\ \hline\hline
SiSTA (base) &51&65.2&58.7&60.4&59.8&62.2&76&58.5&\textbf{82.6}&79.5&\textbf{57.8}&\textbf{69.2}&51.9&74.4&54.8&\textbf{66.4}&\textbf{62.1}&\textbf{77.9}&\textbf{58.2}\\ \hline
SiSTA (prune-zero)&50.5&\textbf{66.3}&\textbf{59.3}&59.4&\textbf{70.9}&65.3&75.6&\textbf{66.8}&80.4&76.4&57.7&64.1&50.3&\textbf{78.7}&\textbf{56.3}&58.7&61.1&73.6&57.1\\ \hline
SiSTA (prune-rewind)&50.3&65.6&55.6&\textbf{61.2}&61&65.4&\textbf{76.5}&60.4&82.3&\textbf{80.2}&56.2&64.6&50.4&76&54.9&61.7&61.8&77&53.9\\ \hline\hline
 \cellcolor[HTML]{E6FFF2} Full target DA &67.6&68.8&65.5&75.6&75.9&90.7&78.6&76.7&88&89.3&61.9&74.2&61.2&86.2&61.3&60.2&67.2&84.9&62.2\\ \hline
\end{tabular}
}
\caption{Performance of \alg~on Domain D (Frost) of the CelebA dataset.}
\label{tab:frost}
\end{table*}

\begin{table*}[]
\renewcommand{\arraystretch}{1.5}
\resizebox{0.99\linewidth}{!}{
\begin{tabular}{|l|c|c|c|c|c|c|c|c|c|c|c|c|c|c|c|c|c|c|c|}
\cline{2-20}
\multicolumn{1}{c|}{}                                 & {\rotatebox[origin=c]{90}{5'o clock shadow}} & \rotatebox[origin=c]{90}{Arched eyebrows} & \rotatebox[origin=c]{90}{Bald} & \rotatebox[origin=c]{90}{Bangs} & \rotatebox[origin=c]{90}{Blond hair} & \rotatebox[origin=c]{90}{Eyeglasses} & \rotatebox[origin=c]{90}{Makeup} & \rotatebox[origin=c]{90}{Cheekbones} & \rotatebox[origin=c]{90}{Gender} & \rotatebox[origin=c]{90}{Mouth open} & \rotatebox[origin=c]{90}{Eyes closed} & \rotatebox[origin=c]{90}{Beard} & \rotatebox[origin=c]{90}{Sideburns} & \rotatebox[origin=c]{90}{Smiling} & \rotatebox[origin=c]{90}{Straight hair} & \rotatebox[origin=c]{90}{Wavy hair} & \rotatebox[origin=c]{90}{Earrings} & \rotatebox[origin=c]{90}{Lipstick} & \rotatebox[origin=c]{90}{Young} \\

\cline{2-20} \hline
 \cellcolor[HTML]{FFE6E6}Source only &50.1&52.5&50.9&59.6&50.2&73.9&51.9&50.1&76.9&66.4&50.1&70.6&57.7&62.9&50.8&61.6&51.2&54&61.8 \\ \hline\hline
MEMO (Augmix) &50&52.7&50&60&50&73.5&51.5&50&77&69.6&50&70.2&58.7&64.4&50.7&61&51.3&54.3&61.7\\ \hline
MEMO (Randconv) &50&52.7&50&60&50&73.5&51.3&50&77&69.7&50&70&58.8&64.7&50.8&60.9&51.3&54.3&61.6\\ \hline\hline
SiSTA (base) &62&64.2&60.9&67.2&79.1&82.6&79.4&65&83.4&77.4&\textbf{68.7}&\textbf{82.1}&\textbf{60.3}&75.4&54.7&75.6&\textbf{67.1}&84.3&66.6\\ \hline
SiSTA (prune-zero)&\textbf{62.4}&63.2&\textbf{61.4}&70.7&\textbf{87.2}&\textbf{87.8}&79.5&\textbf{68.8}&86.2&75.7&67.8&81.4&59.7&79.4&\textbf{57.9}&\textbf{76.1}&64.9&86.4&\textbf{67.4}\\ \hline
SiSTA (prune-rewind)&59.4&\textbf{67.5}&57.5&\textbf{73.6}&79.4&86.2&\textbf{83.3}&65.9&\textbf{86.7}&\textbf{79.3}&66.7&81.2&58.7&\textbf{79.7}&55.8&74.9&67&\textbf{87.1}&65.9\\ \hline\hline
 \cellcolor[HTML]{E6FFF2} Full target DA &76.32&77.68&66.79&82.68&85.69&94.96&84.32&77.87&89.45&83&70.09&85.44&83.71&85.55&62.11&72.93&79.14&84.89& 65.89\\ \hline
\end{tabular}
}
\caption{Performance of \alg~on Domain D (Snow) of the CelebA dataset.}
\label{tab:snow}
\end{table*}

\begin{table*}[]
\renewcommand{\arraystretch}{1.5}
\resizebox{0.99\linewidth}{!}{
\begin{tabular}{|l|c|c|c|c|c|c|c|c|c|c|c|c|c|c|c|c|c|c|c|}
\cline{2-20}
\multicolumn{1}{c|}{}                                 & {\rotatebox[origin=c]{90}{\textcolor{red}{5'o clock shadow}}} & \rotatebox[origin=c]{90}{Arched eyebrows} & \rotatebox[origin=c]{90}{Bald} & \rotatebox[origin=c]{90}{Bangs} & \rotatebox[origin=c]{90}{Blond hair} & \rotatebox[origin=c]{90}{\textcolor{red}{Eyeglasses}} & \rotatebox[origin=c]{90}{Makeup} & \rotatebox[origin=c]{90}{Cheekbones} & \rotatebox[origin=c]{90}{Gender} & \rotatebox[origin=c]{90}{Mouth open} & \rotatebox[origin=c]{90}{\textcolor{red}{Eyes closed}} & \rotatebox[origin=c]{90}{Beard} & \rotatebox[origin=c]{90}{\textcolor{red}{Sideburns}} & \rotatebox[origin=c]{90}{Smiling} & \rotatebox[origin=c]{90}{Straight hair} & \rotatebox[origin=c]{90}{Wavy hair} & \rotatebox[origin=c]{90}{\textcolor{red}{Earrings}} & \rotatebox[origin=c]{90}{Lipstick} & \rotatebox[origin=c]{90}{Young} \\

\cline{2-20} \hline
 \cellcolor[HTML]{FFE6E6}Source only &50&50&50.4&53&53.4&51.5&50&51.2&69.7&54.5&50&58.9&50&59.4&50.5&50.8&50&56.5&61.6 \\ \hline\hline
MEMO (Augmix) &50&50&52.6&51.8&51.9&52.1&50&51.1&68.9&54.5&50&\textbf{58.8}&50&58.3&49.9&50.6&50&55.9&57.3\\ \hline
MEMO (Randconv) &50&50&52.6&51.8&51.4&52.1&50&51.1&69&54.6&50&\textbf{58.8}&50&58.3&49.9&50.6&50&55.7&58.1\\ \hline\hline
SiSTA (base) &50&60.1&54.1&70.3&66.7&50.3&72&65.5&\textbf{83.5}&75.3&50.8&52.5&50.6&74.4&51.7&\textbf{70.6}&51.2&\textbf{77.3}&\textbf{62.5}\\ \hline
SiSTA (prune-zero)&50&\textbf{65.7}&\textbf{58.7}&\textbf{76.4}&\textbf{75.6}&51.1&\textbf{80.1}&\textbf{73.7}&74.2&73.2&50.3&51.1&50.2&\textbf{82.9}&54.9&67.1&52.2&72&57.8\\ \hline
SiSTA (prune-rewind)&50&63.4&55&72.8&72&51&76&67&81.6&\textbf{76.9}&50.5&52.4&50.2&78.5&\textbf{55}&69.8&50.5&76.4&61.5\\ \hline\hline
 \cellcolor[HTML]{E6FFF2} Full target DA &63.2&76.7&65.56&69.31&76.7&92.1&76&76.3&86.4&89.2&58.8&73.1&71.8&87.1&57.1&68.5&73.7&80.5&62.3\\ \hline
\end{tabular}
}
\caption{Performance of \alg~on Domain D (Contrast) of the CelebA dataset.}
\label{tab:contrast}
\end{table*}





\end{document}